\documentclass[11pt]{article}

\usepackage[final]{acl}

\usepackage{times}
\usepackage{latexsym}

\usepackage[T1]{fontenc}
\usepackage[utf8]{inputenc}

\usepackage{microtype}

\usepackage{inconsolata}

\usepackage{graphicx}
\usepackage{microtype}      % microtypography
\usepackage{xcolor}         % colors
\usepackage{enumitem}
\usepackage{multirow}
\usepackage{mathtools}
\usepackage{amssymb}
\usepackage{colortbl}
\usepackage{graphicx}
\usepackage{url}
\usepackage{pgfplots}  
\usepackage{caption}      
\usepackage{subcaption} 
\usepackage{amsmath}
\usepackage{booktabs}
\usepackage{wrapfig}
\usepackage{float}
\usepackage{enumitem}
\makeatletter
\renewcommand*{\@fnsymbol}[1]{\ensuremath{\ifcase#1\or \dagger\or \ddagger
\else\@ctrerr\fi}}
\makeatother

\usepackage{tikz}
\usepackage{pifont}

\newcommand{\posdelta}[1]{\textcolor{green!60!black}{$+#1$}}
\newcommand{\negdelta}[1]{\textcolor{red!80!black}{$-#1$}}

\DeclareRobustCommand{\dist}[1]{%
  \tikz[baseline=(d.base)]{%
    \node[rounded corners=2pt, fill=realmblue!10, draw=realmblue!45,
          line width=0.3pt, inner xsep=3pt, inner ysep=1.4pt,
          font=\small\bfseries\sffamily, text=realmblue!85!black]
         (d) {D#1};%
  }%
}
\usepackage{xspace}
\definecolor{realmblue}{HTML}{0072B2}   % Okabe-Ito blue  (recommended)
\definecolor{realmteal}{HTML}{009E73}   % Okabe-Ito bluish green
\definecolor{realmred}{HTML}{D55E00}    % Okabe-Ito vermillion
\definecolor{realmpurple}{HTML}{6A3D9A}
\newcommand{\ourfont}[1]{\textcolor{realmpurple}{\textbf{\textsf{#1}}}}
\newcommand{\methodfont}[1]{{\small\textsf{#1}}}
\newcommand{\realm}{\ourfont{REALM}\xspace}
\newcommand{\noisy}{\methodfont{Noisy}\xspace}
\newcommand{\oracle}{\methodfont{Oracle}\xspace}
\newcommand{\mv}{\methodfont{MV}\xspace}
\newcommand{\ds}{\methodfont{DS}\xspace}
\newcommand{\crown}{%
  \raisebox{-0.15ex}{\includegraphics[trim=270 346 276 311, clip,
                                      height=1.05em]{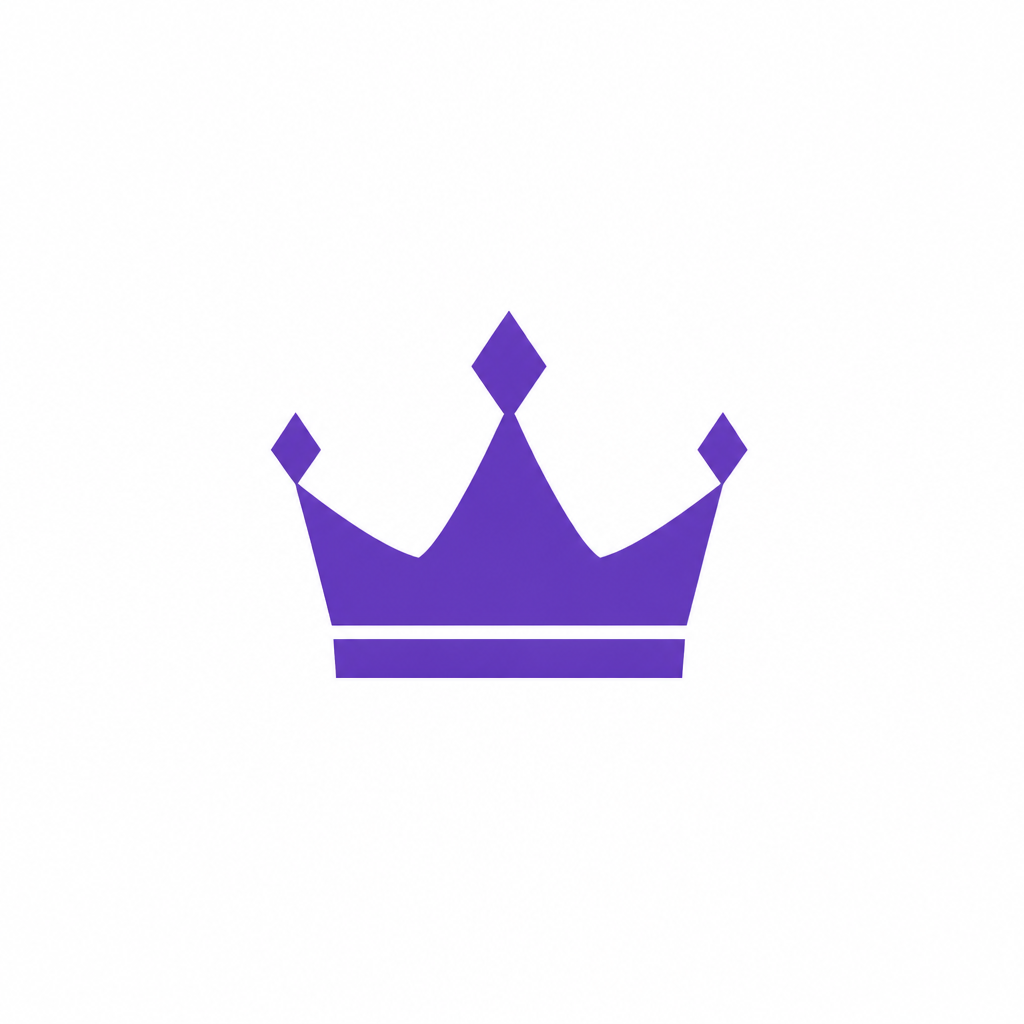}}%
}
\title{\crown\hspace{0.35em}REALM: Reliable Expertise-Aware Language Model Fine-Tuning from Noisy Annotations}

\author{Sajjad Ghiasvand, Mark Beliaev, Mahnoosh Alizadeh, and Ramtin Pedarsani \\
Department of Electrical and Computer Engineering \\
UC Santa Barbara \\
Santa Barbara, CA 93106, USA \\
\texttt{\{sajjad,mbeliaev,alizadeh,ramtin\}@ucsb.edu}}

\begin{document}
\maketitle

\begin{abstract}
Supervised fine-tuning of large language models relies on
human-annotated data, yet annotation pipelines routinely involve
multiple crowdworkers of heterogeneous expertise.
Standard practice aggregates labels via majority vote or simple
averaging, discarding annotator identity and causing the model to
absorb the errors of unreliable annotators into its parameters.
We propose \realm, which jointly learns the model parameters and a
scalar expertise value for each annotator, entirely unsupervised and
requiring nothing beyond annotator identity.
The key idea is to model each observed label as a mixture between
the model's prediction and a uniform random guess, weighted by the
annotator's learned expertise.
\realm applies to any task with a fixed label set, and extends to
multiple tasks via a learned expertise matrix.
On four text-classification datasets with \emph{real} crowdsourced
annotations, \realm is the best method in all $12$ configurations of
the three heterogeneous-annotator datasets, improving on the
strongest applicable baseline, including majority vote and
Dawid--Skene aggregation, by $+2.9$ points on average.
On five question answering benchmarks with simulated noisy labels, it
outperforms naive noisy fine-tuning in $152$ of $162$ configurations,
by $+5.0$ points on average, with gains that grow with model capacity.
The learned expertise additionally recovers annotator reliability
without ever observing it.
Our code is available at
\url{https://github.com/sajjad-ucsb/REALM}.

\end{abstract}

\section{Introduction}

Supervised fine-tuning (SFT) on human-annotated data is a cornerstone
of modern language model development \cite{ouyang2022training,
chung2022scaling, ziegler2019finetuning, bai2022training}.
In practice, annotation pipelines rely on multiple crowdworkers who
vary widely in domain expertise, task familiarity, and attention
\cite{ouyang2022training, bai2022training}.
Despite this, the overwhelming majority of fine-tuning pipelines treat
all annotations as equally reliable, aggregating labels via majority
vote or simple averaging before training \cite{dawid1979maximum,
raykar2010learning}.
This discards information about annotator identity and quality, and
causes the model to absorb the errors of unreliable annotators directly
into its parameters.

The problem is particularly acute in the low-budget crowdsourcing
setting, where a small number of annotators with heterogeneous expertise
label large quantities of data.
When one annotator is highly reliable and two are near-random, a model
trained naively on the combined annotations is learning from noise the
majority of the time.
Recent work has shown that annotation quality matters far more than
quantity for SFT \cite{zhou2023lima}, yet existing methods for handling
label noise \cite{natarajan2013learning, han2018co, patrini2017making}
typically assume a fixed noise transition matrix and require either a
clean validation set or strong assumptions about the noise structure.
Neither is available in the crowdsourcing setting.
The question we address is whether the model can instead identify and
down-weight unreliable annotators automatically, using only the observed
labels and annotator identities, without any additional supervision on
annotator quality.

We propose \textbf{R}eliable \textbf{E}xpertise-\textbf{A}ware
\textbf{L}anguage \textbf{M}odel (\realm) Fine-Tuning, which jointly
learns the model parameters and an expertise value for each annotator.
The key idea is to model each observed label as a mixture of the
model's prediction and a uniform random guess, weighted by the
annotator's expertise.
An annotator with high expertise pushes the model toward their label;
one with low expertise contributes little signal.
Crucially, expertise is never observed: it is recovered entirely
unsupervised from the structure of the training data.
This mixture formulation was introduced in the imitation learning
literature \cite{beliaev2022imitation, beliaev2025inverse} for
unsupervised recovery of demonstrator quality in sequential
decision-making.
We adapt it to LLM fine-tuning by restricting the model's output
distribution to the $K$ label tokens of the task, and extend it to a
multi-task setting where expertise varies across tasks and is encoded
in a learned matrix.
To our knowledge the mixture approach has not previously been applied
to language model training.

We evaluate \realm on four text-classification datasets carrying
\emph{real} crowdsourced annotations \cite{wulczyn2017ex,
rodrigues2013learning, demszky2020goemotions}, whose workers range
from near-chance to near-perfect agreement with the gold label, and on
five multiple-choice question answering benchmarks
\cite{mihaylov2018obqa, clark2018arc, bisk2020piqa, lin2021riddlesense,
jin2019pubmedqa} with simulated annotations, fine-tuning three sizes of
Flan-T5 \cite{chung2022scaling}.
The simulated benchmarks let us vary the expertise distribution, the
noise type, and the annotator pool independently, giving a controlled
account of when and why expertise-aware training helps.

Our main contributions are as follows:
\begin{itemize}
    \item We propose \realm, a mixture-based fine-tuning objective that
    jointly learns model parameters and per-annotator expertise
    entirely unsupervised, requiring no supervision beyond annotator
    identity.

    \item On real crowdsourced annotations, \realm is the best method
    in all $12$ configurations of the three datasets with heterogeneous
    annotators, beating the strongest applicable baseline by $+2.9$
    points on average and $+7.7$ at best.

    \item Under simulated noise, \realm outperforms noisy SFT in $152$
    of $162$ configurations, by $+5.0$ points on average and $+5.5$ on
    the single-task grid, with gains that grow with model capacity and
    persist when the noise violates the model's generative
    assumptions.

    \item The learned expertise values recover annotator reliability
    without ever observing it, providing an interpretable by-product of
    training.
\end{itemize}

% \section{Related Work}
% \input{src/relwork}

\section{Background and Related Work}
\label{sec:related}

\subsection{Learning from Noisy and Crowdsourced Labels}

Inferring ground-truth labels from multiple noisy annotators is
well studied.
\citet{dawid1979maximum} introduced a seminal EM framework that
estimates per-annotator error rates without gold labels, modelling
each annotator by a confusion matrix; \citet{raykar2010learning}
estimate the classifier and annotator parameters jointly, improving
over majority voting; and \citet{whitehill2009vote} add item difficulty
alongside annotator expertise.
These methods infer labels for a fixed classification model, so within
an LLM pipeline they can only act as a preprocessing step: aggregate
first, then fine-tune on the result.
We compare against exactly that pipeline (Section~\ref{sec:exp_setup})
and find that estimating reliability jointly with the model beats
committing to aggregated labels beforehand.
Aggregation also needs several labels per item, and is undefined when
each item is annotated once.

In NLP, annotator disagreement has received growing attention,
particularly for subjective tasks such as toxicity detection and
sentiment analysis \cite{davani2022dealing, gordon2022jury,
plank2022problem, uma2021survey}.
\citet{davani2022dealing} show that multi-task models predicting each
annotator's label separately outperform majority-vote aggregation,
and \citet{gordon2022jury} propose jury learning, which selects a
composition of annotators at inference time to reflect a desired
social perspective.
These works treat disagreement as signal to preserve.
We instead target disagreement that reflects differing annotator
\emph{reliability} with respect to a gold reference, and aim to
recover that reference rather than model the annotator population.

A broader line of work addresses label noise in deep learning
\cite{natarajan2013learning, han2018co, patrini2017making}, and recent
methods extend noise-robust training to LLM fine-tuning
\cite{luo2024robustft, wang2023noiserobust, yadav2023symnoise,
yu2020finetuning}.
These detect and relabel noisy responses at the \emph{instance} level,
typically using an auxiliary model or a clean reference set, and
ignore annotator identity, so they address a different resource
setting from ours.
\realm needs neither a clean reference nor an external checker: it
learns annotator reliability unsupervised during training, and
attributes noise to its source rather than to individual examples.
Robustness to corrupted or adversarially perturbed \emph{inputs}
\citep{goodfellow2015explaining, ghiasvand2025robust} targets
worst-case corruption of the input and is orthogonal.

\subsection{Learning from Suboptimal Demonstrators}

Imitation learning from heterogeneous demonstrators is well studied in
sequential decision-making.
Behavioral cloning \cite{pomerleau1991efficient} treats all
demonstrators as equally expert, degrading the policy when the data
contains suboptimal behavior, and existing remedies require extra
supervision: confidence scores \cite{zhang2021confidence}, a subset of
known-expert demonstrations \cite{yang2021trail}, or environment
dynamics \cite{brown2020better}.

A key insight motivating \realm is that annotator identity alone, with
no quality labels, suffices to recover reliable expertise estimates.
A mixture-based demonstrator model captures this: the observed label
is a convex combination of the optimal prediction and a uniform random
guess, weighted by learned expertise.
Introduced for sequential decision-making, it recovers demonstrator
quality unsupervised while improving the learned policy
\cite{beliaev2022imitation, beliaev2025inverse}.
We adapt it to LLM fine-tuning, where supervision is token-level and
expertise varies across tasks; to our knowledge the mixture approach
has not previously been applied to language model training.

\subsection{Supervised Fine-Tuning and Alignment of Language Models}

Supervised fine-tuning on human-annotated demonstrations is the
standard first step in aligning language models with human intent
\cite{ouyang2022training, chung2022scaling, ziegler2019finetuning,
bai2022training}, and the pipelines producing SFT data rely on
crowdworkers of varying expertise, task familiarity, and attention
\cite{ouyang2022training, bai2022training}.
Most pipelines nevertheless aggregate by majority vote before
training, discarding annotator identity.
Quality matters more than quantity at this stage: a small curated set
can match models trained on orders of magnitude more data
\cite{zhou2023lima}, and instruction data can be generated without
human annotation \cite{wang2023selfinstruct}, which makes automatic
identification of low-quality annotations valuable.

Reinforcement learning from human feedback
\cite{ouyang2022training, stiennon2020learning, bai2022training}
addresses disagreement through preference modelling, but needs
pairwise comparisons rather than direct labels and adds a reward-model
stage with its own noise and bias.
\realm instead operates entirely within SFT: it consumes the same
(prompt, annotator, label) triples and replaces cross-entropy with an
expertise-aware mixture, requiring no preference data, clean reference
set, reward model, or annotation beyond annotator identity.
This makes it a lightweight alternative to standard SFT whenever
training data comes from annotators of heterogeneous quality.

\section{Proposed Algorithm}

\subsection{Problem Setup}
\label{sec:setup}

We consider supervised fine-tuning of a language model on $K$-way
classification with a fixed label set, under noisy labels.
This covers multiple-choice question answering, where the label set is
the answer options of a question, and text classification, where it is
a fixed inventory of categories such as sentiment or toxicity.
Let $\mathcal{Q}$ denote the space of prompts and $\mathcal{A} = \{1, \ldots, K\}$ the label set.
We assume access to a training set of triples $(Q, i, \tilde{Y})$, where $Q \in \mathcal{Q}$ is a prompt, $i \in \{1, \ldots, N\}$ indexes an annotator, and $\tilde{Y} \in \mathcal{A}$ is a noisy observed label.
The ground-truth label $Y^* \in \mathcal{A}$ is never observed during training.

Each annotator has an \textit{expertise level} $\beta_i \in [0, 1]$, interpreted as the probability that their label matches the ground truth.
Standard SFT ignores annotator identity entirely, treating all $(Q, \tilde{Y})$ pairs as equally reliable, thereby absorbing the errors of low-expertise annotators directly into the model.

\subsection{The REALM Objective}
\label{sec:realm_single}

\realm models the observed label as a mixture of the model's
prediction and a uniform random guess, weighted by the annotator's
expertise. Let $\theta$ denote the language model's parameters
and $P_\theta(y \mid Q)$ its predicted probability for label $y$
given prompt $Q$. The observed label distribution is:
\begin{equation}
    P(\tilde{Y} = y \mid Q, i)
    \;=\;
    \beta_i  P_\theta(y \mid Q)
    \;+\;
    \frac{1 - \beta_i}{K}.
    \label{eq:mixture}
\end{equation}
When $\beta_i = 1$ the annotator is a perfect expert; when $\beta_i = 0$ they label uniformly at random.

\textbf{Restricted softmax.}
We require only that each of the $K$ classes be presented as a
lettered option in the prompt and predicted as the corresponding
single answer-letter token (\texttt{a}, \texttt{b}, \texttt{c},
\ldots).
For multiple-choice question answering these are the answer options of
the question; for text classification they are the categories of the
task, so a toxicity prompt lists \texttt{a} toxic, \texttt{b}
neutral, \texttt{c} healthy.
Rather than computing a softmax over the full vocabulary of tens of
thousands of tokens, we restrict the output distribution to these $K$
answer tokens $\{t_1, \ldots, t_K\}$, one per class:
\begin{equation}
    P_\theta(y = k \mid Q)
    \;=\;
    \frac{\exp(z_{t_k})}{\sum_{k'=1}^{K} \exp(z_{t_{k'}})},
    \label{eq:restricted_softmax}
\end{equation}
where $z_v$ denotes the logit for vocabulary token $v$ at the answer
position in the generated sequence.
This serves two purposes: it ensures $P_\theta$ is a proper
distribution over $\mathcal{A}$ rather than being diluted across
irrelevant tokens, and it aligns the model's output space with the
uniform term $1/K$ in the mixture objective of
Equation~\ref{eq:mixture}, so that both components operate over the
same $K$-class distribution.

\textbf{Learning expertise.}
We parameterize each annotator's expertise as $\beta_i = \sigma(\omega_i)$, where $\omega_i \in \mathbb{R}$ is a scalar trained jointly with the model parameters $\theta$.
The training objective is the negative log-likelihood of the observed labels under the mixture model:
%
% \begin{equation}
%     \mathcal{L}(\theta, \boldsymbol{\omega})
%     \;=\;
%     -\sum_{(Q,\, i,\, \tilde{Y})}
%     \log\!\left[
%         \sigma(\omega_i) \cdot P_\theta(\tilde{Y} \mid Q)
%         \;+\;
%         \frac{1 - \sigma(\omega_i)}{K}
%     \right],
%     \label{eq:loss}
% \end{equation}

\begin{align}
    \mathcal{L}(\theta, \boldsymbol{\omega})
    &= -\sum_{(Q,\, i,\, \tilde{Y})}
    \log\Bigl[
        \sigma(\omega_i) \cdot P_\theta(\tilde{Y} \mid Q) \notag\\
    &\qquad\qquad+\; \frac{1 - \sigma(\omega_i)}{K}
    \Bigr].
    \label{eq:loss}
\end{align}
where $\boldsymbol{\omega} = \{\omega_i\}_{i=1}^{N}$.
We optimise $\theta$ and $\boldsymbol{\omega}$ jointly using AdamW with separate learning rates, as the expertise scalars typically benefit from a larger step size than $\theta$.
At convergence, $\sigma(\omega_i)$ is an interpretable, unsupervised
estimate of annotator $i$'s reliability, recovered accurately for
reliable annotators and shrunk toward zero for those whose labels
carry little signal (Section~\ref{sec:expertise}).

\textbf{Relationship to standard SFT.}
When $\omega_i \to +\infty$ for all $i$, all $\beta_i \to 1$ and $\mathcal{L}$ reduces to the standard cross-entropy loss on the restricted softmax.
\realm therefore strictly generalises SFT, recovering it in the high-expertise limit and progressively down-weighting annotator errors as expertise falls.

\subsection{Extension of REALM to the Multi-Task Setting}
\label{sec:realm_multi}

We extend \realm to a setting with $M$ tasks (datasets), each with its
own label set $\mathcal{A}^{(m)}$ of size $K_m$.
Let $f^*\!: \mathcal{Q} \to \{1, \ldots, M\}$ be the ground-truth task
assignment and write $T = f^*(Q)$.
In our experiments, $f^*(Q)$ is always observed since we know which
dataset each prompt originates from.
Each annotator $i$ is assigned a \textit{skill vector}
$\boldsymbol{\omega}_i \in \mathbb{R}^M$, generalising the scalar
$\omega_i$ from Section~\ref{sec:realm_single} to one entry per task,
where $\omega_{i,m}$ captures the reliability of annotator $i$ on
task $m$.
Stacking these vectors as rows defines a learned $N \times M$ expertise
matrix $\boldsymbol{\Omega}$, where $\Omega_{i,m} = \omega_{i,m}$,
and the expertise of annotator $i$ on a prompt from task $T$ is
$\sigma(\Omega_{i,T})$.
The multi-task training objective is:
%
% \begin{equation}
%     \mathcal{L}(\theta, \boldsymbol{\Omega})
%     \;=\;
%     -\sum_{(Q,\, i,\, \tilde{Y})}
%     \log\!\left[
%         \sigma\!\left(\Omega_{i,\, f^*(Q)}\right)
%         \cdot P_\theta(\tilde{Y} \mid Q)
%         \;+\;
%         \frac{1 - \sigma\!\left(\Omega_{i,\, f^*(Q)}\right)}{K_{f^*(Q)}}
%     \right].
%     \label{eq:multi_loss}
% \end{equation}

% \begin{align}
%     \mathcal{L}(\theta, \boldsymbol{\Omega})
%     &= -\sum_{(Q,\, i,\, \tilde{Y})}
%     \log\Bigl[
%         \sigma\!\left(\Omega_{i,\, f^*(Q)}\right)
%         \cdot P_\theta(\tilde{Y} \mid Q) \notag\\
%     &\qquad\qquad+\; \frac{1 - \sigma\!\left(\Omega_{i,\, f^*(Q)}\right)}
%                          {K_{f^*(Q)}}
%     \Bigr].
%     \label{eq:multi_loss}
% \end{align}

\begin{align}
    \mathcal{L}(\theta, \boldsymbol{\Omega})
    &= -\sum_{(Q,\, i,\, \tilde{Y})}
    \log\Bigl[
        \sigma\!\left(\Omega_{i,\, T}\right)
        \cdot P_\theta(\tilde{Y} \mid Q) \notag\\
    &\qquad\qquad+\; \frac{1 - \sigma\!\left(\Omega_{i,\, T}\right)}
                         {K_{T}}
    \Bigr].
    \label{eq:multi_loss}
\end{align}
where $K_T = K_{f^*(Q)}$ denotes the number of answer choices for task $T$. In all settings the true matrix $\boldsymbol{\Omega}$ is never provided; the model learns it entirely from the observed triples $(Q, i, \tilde{Y})$.

\paragraph{Per-task restricted softmax.}
Because tasks differ in their number of classes, we construct a
separate set of label token ids per task.
Within each minibatch, examples are grouped by task so that the correct
$K_m$ is used for both the restricted softmax and the uniform term
$1/K_m$, ensuring that both components of the objective operate over
the same $K_m$-class distribution.

\section{Empirical Results}

\subsection{Experimental Setup}
\label{sec:exp_setup}

\textbf{Datasets with Real Annotations.}
Our primary evaluation uses four text-classification datasets that
release the individual label of \emph{every} annotator alongside a
gold-standard test set, so that annotator heterogeneity is measured
rather than simulated.
\textbf{WikiAttack} \cite{wulczyn2017ex} labels Wikipedia talk-page
comments as a personal attack or not ($K{=}2$) using $4{,}048$ crowd
workers whose agreement with the gold label ranges from $0.09$ to
$1.00$.
\textbf{WikiToxicity} \cite{wulczyn2017ex} labels the same source of
comments as toxic, neutral, or healthy ($K{=}3$) using $4{,}301$
workers with agreement between $0.21$ and $0.94$.
\textbf{SentPolarity} \cite{rodrigues2013learning} is the standard
learning-from-crowds benchmark of movie-review sentences ($K{=}2$)
annotated by $203$ crowd workers with accuracy between $0.20$ and
$0.98$.
\textbf{GoEmotions} \cite{demszky2020goemotions} maps Reddit comments
to the six Ekman emotions plus neutral ($K{=}7$) using $82$ trained raters of
homogeneous quality; we include it as a \emph{control} on which
expertise modelling should bring no benefit.
For each dataset we construct a training pool of the $m$ lowest- and
$m$ highest-agreement annotators (``contrast-$m$'', with $m{=}150$,
$150$, $20$, and $10$ respectively), which reproduces the wide
expertise spread that motivates \realm while keeping the pool
tractable.

\textbf{Datasets with Simulated Annotations.}
To study the effect of expertise structure and noise type under full
control, we additionally simulate annotations on five multiple-choice
question answering benchmarks.
For commonsense reasoning: OpenBookQA \cite{mihaylov2018obqa} (OBQA)
and ARC \cite{clark2018arc}, each with four answer choices ($K{=}4$);
PIQA \cite{bisk2020piqa}, a physical intuition benchmark with two
choices ($K{=}2$); and RiddleSense \cite{lin2021riddlesense}, a
commonsense riddle dataset with five choices ($K{=}5$).
For biomedical reasoning: PubMedQA \cite{jin2019pubmedqa}, a
yes/no/maybe question answering dataset with three choices ($K{=}3$).
These datasets span a range of task difficulties, domain types, and
answer-set sizes, making them well-suited for studying how annotation
noise and expertise structure interact with model performance.
In the multi-task setting we combine three datasets;
single-task experiments are conducted on each dataset independently.

\textbf{Annotation Regimes.}
We consider two regimes differing in how many annotators label each
example.
In the \textbf{sparse} regime, every example carries exactly one
annotation, so the dataset size matches the original and each
annotator covers a disjoint portion of the data.
This is the common crowdsourcing scenario under a fixed budget,
and it is the regime in which label aggregation is
\emph{impossible}: with one label per example there is nothing to
aggregate.
In the \textbf{shared} regime, every annotator labels every example,
which multiplies both the amount of supervision and the proportion of
corrupted labels by the pool size.
Unless stated otherwise, simulated experiments use the
sparse regime, and real-annotation experiments report both.

\begin{table*}[t]
\centering
\caption{Test accuracy (\%) on the datasets with real crowd
annotations, averaged over three seeds.
\mv and \ds need several labels per item and are inapplicable
in the sparse regime; $\Delta$ is \realm minus the best applicable
baseline.}
\label{tab:results_real}
\resizebox{0.65\textwidth}{!}{%
\begin{tabular}{|l|l|ccc|ccccc|}
\hline
\multirow{2}{*}{\textbf{Dataset}} & \multirow{2}{*}{\textbf{Size}}
& \multicolumn{3}{c|}{\textbf{Sparse} (one label / item)}
& \multicolumn{5}{c|}{\textbf{Shared} (all labels kept)} \\
\cline{3-10}
& & \realm & \noisy & $\Delta$
  & \realm & \noisy & \mv & \ds & $\Delta$ \\
\hline
\multirow{2}{*}{WikiAttack}
& Base  & $\mathbf{92.4}$ & $84.7$ & \posdelta{7.7} & $\mathbf{93.5}$ & $84.6$ & $84.9$ & $91.6$ & \posdelta{1.9} \\
& Large & $\mathbf{89.2}$ & $85.7$ & \posdelta{3.5} & $\mathbf{92.9}$ & $84.9$ & $85.2$ & $91.4$ & \posdelta{1.5} \\
\hline
\multirow{2}{*}{WikiToxicity}
& Base  & $\mathbf{67.8}$ & $63.7$ & \posdelta{4.1} & $\mathbf{73.1}$ & $70.8$ & $66.6$ & $69.9$ & \posdelta{2.3} \\
& Large & $\mathbf{69.7}$ & $66.4$ & \posdelta{3.3} & $\mathbf{73.8}$ & $71.5$ & $68.9$ & $70.4$ & \posdelta{2.3} \\
\hline
\multirow{2}{*}{SentPolarity}
& Base  & $\mathbf{87.0}$ & $84.2$ & \posdelta{2.8} & $\mathbf{89.3}$ & $88.2$ & $86.7$ & $88.6$ & \posdelta{0.7} \\
& Large & $\mathbf{89.6}$ & $85.7$ & \posdelta{3.9} & $\mathbf{91.7}$ & $90.4$ & $88.3$ & $90.7$ & \posdelta{1.0} \\
\hline
\multirow{2}{*}{GoEmotions}
& Base  & $\mathbf{62.5}$ & $62.1$ & \posdelta{0.4} & $61.5$ & $61.3$ & $\mathbf{61.6}$ & $60.6$ & \negdelta{0.1} \\
& Large & $\mathbf{63.3}$ & $63.2$ & \posdelta{0.1} & $62.7$ & $\mathbf{62.9}$ & $62.7$ & $61.3$ & \negdelta{0.2} \\
\hline
\end{tabular}%
}
\end{table*}

\begin{table*}[t]
\centering
\caption{Test accuracy (\%) under uniform noise (mean $\pm$ std over five runs).
\textbf{Bold} indicates the better method and $\Delta = \realm - \noisy$.}
\label{tab:results_uniform}
\resizebox{0.9\textwidth}{!}{%
\begin{tabular}{|l|l|c|ccc|ccc|ccc|}
\hline
\multirow{2}{*}{\textbf{Size}} & \multirow{2}{*}{\textbf{Dataset}} & \multirow{2}{*}{\textbf{Oracle}}
& \multicolumn{3}{c|}{\dist{1}}
& \multicolumn{3}{c|}{\dist{2}}
& \multicolumn{3}{c|}{\dist{3}} \\
\cline{4-12}
& & & \realm & \noisy & $\Delta$
      & \realm & \noisy & $\Delta$
      & \realm & \noisy & $\Delta$ \\
\hline
\multirow{6}{*}{Small}
& ARC      & 29.44 & $\mathbf{27.35}_{\pm 0.63}$ & $26.68_{\pm 1.16}$ & \posdelta{0.67} & $\mathbf{29.05}_{\pm 0.47}$ & $27.00_{\pm 0.87}$ & \posdelta{2.05} & $\mathbf{26.70}_{\pm 0.51}$ & $26.64_{\pm 1.40}$ & \posdelta{0.06} \\
% & BioASQ   & 81.30 & $\mathbf{55.45}_{\pm 2.59}$ & $27.15_{\pm 1.90}$ & \posdelta{28.30} & $60.98_{\pm 0.89}$ & $\mathbf{72.36}_{\pm 4.66}$ & \negdelta{11.38} & $\mathbf{53.98}_{\pm 2.22}$ & $51.38_{\pm 7.97}$ & \posdelta{2.60} \\
& OBQA     & 44.20 & $\mathbf{39.44}_{\pm 1.26}$ & $30.28_{\pm 2.07}$ & \posdelta{9.16} & $\mathbf{44.28}_{\pm 0.39}$ & $39.52_{\pm 0.94}$ & \posdelta{4.76} & $\mathbf{42.52}_{\pm 0.45}$ & $35.80_{\pm 2.04}$ & \posdelta{6.72} \\
& PIQA     & 52.88 & $\mathbf{52.14}_{\pm 1.56}$ & $51.08_{\pm 1.35}$ & \posdelta{1.06} & $\mathbf{51.60}_{\pm 1.54}$ & $51.12_{\pm 0.86}$ & \posdelta{0.48} & $\mathbf{52.88}_{\pm 1.18}$ & $50.36_{\pm 0.60}$ & \posdelta{2.52} \\
& PubMedQA & 58.00 & $\mathbf{50.60}_{\pm 1.22}$ & $35.95_{\pm 6.23}$ & \posdelta{14.65} & $\mathbf{53.05}_{\pm 1.88}$ & $50.35_{\pm 3.00}$ & \posdelta{2.70} & $\mathbf{50.60}_{\pm 1.97}$ & $46.55_{\pm 3.17}$ & \posdelta{4.05} \\
& Riddle   & 39.02 & $\mathbf{31.49}_{\pm 1.30}$ & $23.80_{\pm 0.86}$ & \posdelta{7.69} & $\mathbf{34.71}_{\pm 1.14}$ & $31.25_{\pm 1.11}$ & \posdelta{3.46} & $\mathbf{32.16}_{\pm 1.44}$ & $29.69_{\pm 0.69}$ & \posdelta{2.47} \\
\cline{2-12}
& Avg & 44.71 & $\mathbf{40.20}_{\pm 1.19}$ & $33.56_{\pm 2.33}$ & \posdelta{6.65} & $\mathbf{42.54}_{\pm 1.08}$ & $39.85_{\pm 1.36}$ & \posdelta{2.69} & $\mathbf{40.97}_{\pm 1.11}$ & $37.81_{\pm 1.58}$ & \posdelta{3.16} \\
\hline
\multirow{6}{*}{Base}
& ARC      & 47.47 & $\mathbf{43.97}_{\pm 0.77}$ & $40.55_{\pm 1.38}$ & \posdelta{3.42} & $\mathbf{45.49}_{\pm 0.80}$ & $44.31_{\pm 1.01}$ & \posdelta{1.18} & $\mathbf{43.78}_{\pm 0.60}$ & $42.15_{\pm 0.39}$ & \posdelta{1.63} \\
% & BioASQ   & 87.80 & $\mathbf{67.15}_{\pm 0.83}$ & $21.95_{\pm 3.13}$ & \posdelta{45.20} & $66.34_{\pm 2.28}$ & $\mathbf{72.68}_{\pm 3.97}$ & \negdelta{6.34} & $\mathbf{66.99}_{\pm 2.09}$ & $58.54_{\pm 7.88}$ & \posdelta{8.45} \\
& OBQA     & 59.60 & $\mathbf{57.88}_{\pm 0.24}$ & $49.12_{\pm 1.66}$ & \posdelta{8.76} & $\mathbf{58.68}_{\pm 0.81}$ & $56.84_{\pm 0.64}$ & \posdelta{1.84} & $\mathbf{57.96}_{\pm 0.54}$ & $55.00_{\pm 1.28}$ & \posdelta{2.96} \\
& PIQA     & 67.68 & $\mathbf{65.70}_{\pm 0.67}$ & $46.31_{\pm 2.37}$ & \posdelta{19.39} & $\mathbf{66.55}_{\pm 0.52}$ & $63.33_{\pm 1.97}$ & \posdelta{3.22} & $\mathbf{65.79}_{\pm 0.78}$ & $51.47_{\pm 1.23}$ & \posdelta{14.32} \\
& PubMedQA & 61.00 & $\mathbf{51.80}_{\pm 0.97}$ & $40.60_{\pm 2.96}$ & \posdelta{11.20} & $\mathbf{54.00}_{\pm 2.19}$ & $53.20_{\pm 2.15}$ & \posdelta{0.80} & $\mathbf{49.15}_{\pm 1.85}$ & $44.80_{\pm 3.18}$ & \posdelta{4.35} \\
& Riddle   & 54.12 & $\mathbf{46.04}_{\pm 2.07}$ & $36.20_{\pm 2.22}$ & \posdelta{9.84} & $\mathbf{49.61}_{\pm 1.12}$ & $45.84_{\pm 1.03}$ & \posdelta{3.77} & $\mathbf{46.63}_{\pm 2.01}$ & $42.00_{\pm 1.88}$ & \posdelta{4.63} \\
\cline{2-12}
& Avg & 57.97 & $\mathbf{53.08}_{\pm 0.94}$ & $42.56_{\pm 2.12}$ & \posdelta{10.52} & $\mathbf{54.87}_{\pm 1.09}$ & $52.70_{\pm 1.36}$ & \posdelta{2.16} & $\mathbf{52.66}_{\pm 1.16}$ & $47.08_{\pm 1.59}$ & \posdelta{5.58} \\
\hline
\multirow{6}{*}{Large}
& ARC      & 64.81 & $\mathbf{57.56}_{\pm 0.80}$ & $53.68_{\pm 1.19}$ & \posdelta{3.88} & $\mathbf{61.51}_{\pm 0.92}$ & $57.96_{\pm 0.88}$ & \posdelta{3.55} & $\mathbf{59.59}_{\pm 0.64}$ & $56.14_{\pm 1.07}$ & \posdelta{3.45} \\
% & BioASQ   & 86.99 & $\mathbf{68.29}_{\pm 2.82}$ & $23.58_{\pm 4.15}$ & \posdelta{44.71} & $67.48_{\pm 1.54}$ & $\mathbf{76.75}_{\pm 1.51}$ & \negdelta{9.27} & $\mathbf{67.32}_{\pm 2.02}$ & $60.16_{\pm 8.46}$ & \posdelta{7.16} \\
& OBQA     & 71.60 & $\mathbf{69.68}_{\pm 0.99}$ & $61.36_{\pm 0.99}$ & \posdelta{8.32} & $\mathbf{70.68}_{\pm 0.43}$ & $68.80_{\pm 0.46}$ & \posdelta{1.88} & $\mathbf{70.12}_{\pm 1.24}$ & $65.68_{\pm 0.52}$ & \posdelta{4.44} \\
& PIQA     & 77.80 & $\mathbf{76.45}_{\pm 0.31}$ & $26.27_{\pm 0.88}$ & \posdelta{50.18} & $\mathbf{77.67}_{\pm 0.64}$ & $75.10_{\pm 0.23}$ & \posdelta{2.57} & $\mathbf{76.50}_{\pm 0.28}$ & $51.99_{\pm 3.98}$ & \posdelta{24.51} \\
& PubMedQA & 71.25 & $\mathbf{57.65}_{\pm 2.78}$ & $43.85_{\pm 3.11}$ & \posdelta{13.80} & $\mathbf{60.55}_{\pm 1.93}$ & $56.00_{\pm 1.15}$ & \posdelta{4.55} & $\mathbf{53.45}_{\pm 2.09}$ & $48.05_{\pm 3.08}$ & \posdelta{5.40} \\
& Riddle   & 67.84 & $\mathbf{60.43}_{\pm 1.25}$ & $50.35_{\pm 2.16}$ & \posdelta{10.08} & $\mathbf{65.45}_{\pm 0.57}$ & $61.49_{\pm 1.24}$ & \posdelta{3.96} & $\mathbf{62.31}_{\pm 1.65}$ & $57.06_{\pm 2.38}$ & \posdelta{5.25} \\
\cline{2-12}
& Avg & 70.66 & $\mathbf{64.35}_{\pm 1.23}$ & $47.10_{\pm 1.67}$ & \posdelta{17.25} & $\mathbf{67.17}_{\pm 0.90}$ & $63.87_{\pm 0.79}$ & \posdelta{3.30} & $\mathbf{64.39}_{\pm 1.18}$ & $55.78_{\pm 2.21}$ & \posdelta{8.61} \\
\hline
\end{tabular}%
}
\end{table*}

\begin{table*}[t]
\centering
\caption{Test accuracy (\%) under \dist{1} across noise types, datasets, and model sizes (mean $\pm$ std over five runs). \textbf{Bold} indicates the better method and $\Delta = \realm - \noisy$.}
\label{tab:results_noise_types}
\resizebox{0.9\textwidth}{!}{%
\begin{tabular}{|l|l|c|ccc|ccc|ccc|}
\hline
\multirow{2}{*}{\textbf{Size}} & \multirow{2}{*}{\textbf{Dataset}} & \multirow{2}{*}{\textbf{Oracle}}
& \multicolumn{3}{c|}{Uniform}
& \multicolumn{3}{c|}{Asymmetric}
& \multicolumn{3}{c|}{Systematic} \\
\cline{4-12}
& & & \realm & \noisy & $\Delta$
      & \realm & \noisy & $\Delta$
      & \realm & \noisy & $\Delta$ \\
\hline
\multirow{6}{*}{Small}
& ARC      & 29.44 & $\mathbf{27.35}_{\pm 0.63}$ & $26.68_{\pm 1.16}$ & \posdelta{0.67} & $\mathbf{27.04}_{\pm 1.74}$ & $26.97_{\pm 1.33}$ & \posdelta{0.07} & $\mathbf{26.51}_{\pm 1.15}$ & $25.91_{\pm 0.73}$ & \posdelta{0.60} \\
% & BioASQ   & 81.30 & $\mathbf{55.45}_{\pm 2.59}$ & $27.15_{\pm 1.90}$ & \posdelta{28.30} & $\mathbf{53.17}_{\pm 3.97}$ & $27.15_{\pm 4.07}$ & \posdelta{26.02} & $62.60_{\pm 2.06}$ & $\mathbf{70.57}_{\pm 2.79}$ & \negdelta{7.97} \\
& OBQA     & 44.20 & $\mathbf{39.44}_{\pm 1.26}$ & $30.28_{\pm 2.07}$ & \posdelta{9.16} & $\mathbf{37.84}_{\pm 2.69}$ & $31.64_{\pm 2.28}$ & \posdelta{6.20} & $\mathbf{40.80}_{\pm 1.53}$ & $27.44_{\pm 3.43}$ & \posdelta{13.36} \\
& PIQA     & 52.88 & $\mathbf{52.14}_{\pm 1.56}$ & $51.08_{\pm 1.35}$ & \posdelta{1.06} & $\mathbf{51.06}_{\pm 0.96}$ & $50.16_{\pm 0.67}$ & \posdelta{0.90} & $\mathbf{54.41}_{\pm 1.23}$ & $50.84_{\pm 0.92}$ & \posdelta{3.57} \\
& PubMedQA & 58.00 & $\mathbf{50.60}_{\pm 1.22}$ & $35.95_{\pm 6.23}$ & \posdelta{14.65} & $\mathbf{48.80}_{\pm 2.38}$ & $28.80_{\pm 4.19}$ & \posdelta{20.00} & $53.70_{\pm 1.51}$ & $\mathbf{54.20}_{\pm 0.90}$ & \negdelta{0.50} \\
& Riddle   & 39.02 & $\mathbf{31.49}_{\pm 1.30}$ & $23.80_{\pm 0.86}$ & \posdelta{7.69} & $\mathbf{31.45}_{\pm 1.07}$ & $23.53_{\pm 0.89}$ & \posdelta{7.92} & $\mathbf{31.88}_{\pm 1.09}$ & $21.96_{\pm 1.53}$ & \posdelta{9.92} \\
\cline{2-12}
& Avg & 44.71 & $\textbf{40.20}_{\pm 1.19}$ & $33.56_{\pm 2.33}$ & \posdelta{6.65} & $\textbf{39.24}_{\pm 1.77}$ & $32.22_{\pm 1.87}$ & \textbf{\posdelta{7.02}} & $\textbf{41.46}_{\pm 1.30}$ & $36.07_{\pm 1.50}$ & \posdelta{5.39} \\
\hline
\multirow{6}{*}{Base}
& ARC      & 47.47 & $\mathbf{43.97}_{\pm 0.77}$ & $40.55_{\pm 1.38}$ & \posdelta{3.42} & $\mathbf{42.76}_{\pm 0.54}$ & $40.74_{\pm 0.81}$ & \posdelta{2.02} & $\mathbf{43.14}_{\pm 0.51}$ & $41.77_{\pm 0.61}$ & \posdelta{1.37} \\
% & BioASQ   & 87.80 & $\mathbf{67.15}_{\pm 0.83}$ & $21.95_{\pm 3.13}$ & \posdelta{45.20} & $\mathbf{65.85}_{\pm 1.36}$ & $24.88_{\pm 3.43}$ & \posdelta{40.97} & $68.29_{\pm 1.26}$ & $\mathbf{73.17}_{\pm 3.25}$ & \negdelta{4.88} \\
& OBQA     & 59.60 & $\mathbf{57.88}_{\pm 0.24}$ & $49.12_{\pm 1.66}$ & \posdelta{8.76} & $\mathbf{57.48}_{\pm 0.39}$ & $49.84_{\pm 1.49}$ & \posdelta{7.64} & $\mathbf{57.52}_{\pm 0.68}$ & $49.00_{\pm 1.41}$ & \posdelta{8.52} \\
& PIQA     & 67.68 & $\mathbf{65.70}_{\pm 0.67}$ & $46.31_{\pm 2.37}$ & \posdelta{19.39} & $\mathbf{66.22}_{\pm 0.34}$ & $46.31_{\pm 1.79}$ & \posdelta{19.91} & $\mathbf{66.92}_{\pm 0.42}$ & $64.87_{\pm 0.97}$ & \posdelta{2.05} \\
& PubMedQA & 61.00 & $\mathbf{51.80}_{\pm 0.97}$ & $40.60_{\pm 2.96}$ & \posdelta{11.20} & $\mathbf{50.70}_{\pm 1.58}$ & $36.35_{\pm 0.75}$ & \posdelta{14.35} & $\mathbf{55.30}_{\pm 1.18}$ & $53.10_{\pm 1.62}$ & \posdelta{2.20} \\
& Riddle   & 54.12 & $\mathbf{46.04}_{\pm 2.07}$ & $36.20_{\pm 2.22}$ & \posdelta{9.84} & $\mathbf{45.76}_{\pm 1.41}$ & $37.14_{\pm 2.05}$ & \posdelta{8.62} & $\mathbf{46.86}_{\pm 1.75}$ & $35.37_{\pm 2.32}$ & \posdelta{11.49} \\
\cline{2-12}
& Avg & 57.97 & $\textbf{53.08}_{\pm 0.94}$ & $42.56_{\pm 2.12}$ & \posdelta{10.52} & $\textbf{52.58}_{\pm 0.85}$ & $42.08_{\pm 1.38}$ & \textbf{\posdelta{10.50}} & $\textbf{53.95}_{\pm 0.91}$ & $48.82_{\pm 1.39}$ & \posdelta{5.13} \\
\hline
\multirow{6}{*}{Large}
& ARC      & 64.81 & $\mathbf{57.56}_{\pm 0.80}$ & $53.68_{\pm 1.19}$ & \posdelta{3.88} & $\mathbf{57.68}_{\pm 0.75}$ & $53.01_{\pm 0.29}$ & \posdelta{4.67} & $\mathbf{57.48}_{\pm 0.81}$ & $54.99_{\pm 0.38}$ & \posdelta{2.49} \\
% & BioASQ   & 86.99 & $\mathbf{68.29}_{\pm 2.82}$ & $23.58_{\pm 4.15}$ & \posdelta{44.71} & $\mathbf{67.64}_{\pm 2.43}$ & $23.09_{\pm 5.72}$ & \posdelta{44.55} & $67.80_{\pm 1.10}$ & $\mathbf{80.49}_{\pm 3.13}$ & \negdelta{12.69} \\
& OBQA     & 71.60 & $\mathbf{69.68}_{\pm 0.99}$ & $61.36_{\pm 0.99}$ & \posdelta{8.32} & $\mathbf{69.04}_{\pm 0.75}$ & $57.84_{\pm 1.47}$ & \posdelta{11.20} & $\mathbf{69.60}_{\pm 1.00}$ & $61.88_{\pm 1.18}$ & \posdelta{7.72} \\
& PIQA     & 77.80 & $\mathbf{76.45}_{\pm 0.31}$ & $26.27_{\pm 0.88}$ & \posdelta{50.18} & $\mathbf{77.06}_{\pm 0.27}$ & $27.62_{\pm 1.63}$ & \posdelta{49.44} & $\mathbf{77.67}_{\pm 0.63}$ & $75.60_{\pm 1.04}$ & \posdelta{2.07} \\
& PubMedQA & 71.25 & $\mathbf{57.65}_{\pm 2.78}$ & $43.85_{\pm 3.11}$ & \posdelta{13.80} & $\mathbf{58.90}_{\pm 2.37}$ & $35.95_{\pm 1.98}$ & \posdelta{22.95} & $\mathbf{62.10}_{\pm 1.03}$ & $61.05_{\pm 2.65}$ & \posdelta{1.05} \\
& Riddle   & 67.84 & $\mathbf{60.43}_{\pm 1.25}$ & $50.35_{\pm 2.16}$ & \posdelta{10.08} & $\mathbf{59.69}_{\pm 1.32}$ & $50.75_{\pm 3.92}$ & \posdelta{8.94} & $\mathbf{60.55}_{\pm 0.96}$ & $50.08_{\pm 1.54}$ & \posdelta{10.47} \\
\cline{2-12}
& Avg & 70.66 & $\textbf{64.35}_{\pm 1.23}$ & $47.10_{\pm 1.67}$ & \posdelta{17.25} & $\textbf{64.47}_{\pm 1.09}$ & $45.03_{\pm 1.86}$ & \textbf{\posdelta{19.44}} & $\textbf{65.48}_{\pm 0.89}$ & $60.72_{\pm 1.36}$ & \posdelta{4.76} \\
\hline
\end{tabular}%
}
\end{table*}

\textbf{Simulated Annotation Protocol.}
Simulated labels use $N{=}3$ annotators, each emitting the true label
with probability $\beta_i$ and otherwise erring according to one of
the noise types below.
We evaluate three expertise distributions that vary how many
annotators are reliable: \dist{1}
($\beta{=}[0.9, 0.1, 0.1]$), a single reliable annotator outnumbered
by two near-random ones; \dist{2}
($\beta{=}[0.9, 0.9, 0.1]$), a reliable majority; and
\dist{3} ($\beta{=}[0.9, 0.5, 0.1]$), expertise
spread uniformly from high to low.
We cross these with three \emph{noise types}, which control how an
erroneous label is distributed over the $K{-}1$ wrong classes:
\textbf{uniform}, drawn uniformly at random, the channel assumed by
the mixture objective; \textbf{asymmetric}, concentrated on the
cyclically adjacent class, which receives $c{=}4$ times the weight of
the remaining wrong classes; and \textbf{systematic}, a deterministic
annotator-specific offset that makes errors fully predictable from
annotator identity and the true label.
The latter two preserve each annotator's marginal accuracy $\beta_i$
but violate \realm's generative assumption, and therefore measure
robustness to misspecification.
We additionally evaluate a \textbf{multi-task} setting in which three
datasets are trained jointly with $M{=}3$ tasks and a square expertise
matrix $\boldsymbol{\Omega} \in \mathbb{R}^{3 \times 3}$, under three
structured patterns of per-task expertise.
Appendix~\ref{app:simulation} gives the full specification of the
expertise distributions, noise channels, and multi-task
configurations.

\textbf{Baselines.}
We compare \realm against four baselines, all of which receive exactly
the same annotations.
\oracle SFT trains on ground-truth labels and upper-bounds the
achievable accuracy; it is unattainable in practice, since true labels
are unobserved during training.
\noisy SFT trains on the noisy labels with standard
cross-entropy, discarding the annotator index $i$ entirely, and
reflects the standard pipeline used when annotations are collected
without quality control.
It is the exact ablation of our objective: the only difference from
\realm is whether the annotator-aware mixture of
Equation~\ref{eq:loss} replaces cross-entropy.
\mv{}\,+\,SFT and \ds{}\,+\,SFT \cite{dawid1979maximum}
instead collapse the several labels of an example into one before
training, by majority vote and by expectation-maximised per-annotator
confusion matrices respectively, and then run standard SFT on the
aggregated labels; both require multiple labels per example and are
therefore undefined in the sparse regime.
Appendix~\ref{app:baselines} specifies the input and target format of
each method and the inference procedure, which is identical
throughout.

\textbf{Implementation Details.}
We fine-tune three sizes of Flan-T5 \cite{chung2022scaling}: small
(80M), base (250M), and large (780M parameters), using AdamW with a
cosine schedule, a model learning rate of $5{\times}10^{-5}$, and an
effective batch size of $64$.
Simulated-annotation experiments train for $200$ steps and
real-annotation experiments for $1{,}000$--$3{,}000$ steps, which
correspond to between roughly one and forty epochs depending on the
size of the training split; Appendix~\ref{app:impl} reports the
per-dataset budgets.
\realm's expertise scalars $\boldsymbol{\omega}$ are optimised with a
separate learning rate, selected from $\{0.1, 0.5, 1.0\}$ for the
simulated-annotation experiments and from $\{0.005, 0.01, 0.1\}$ for
the real-annotation ones, in both cases on a single seed and reused
unchanged for all others.
These low-dimensional parameters benefit from faster adaptation than
$\theta$; this is \realm's only tuned hyperparameter, and
Appendix~\ref{app:lr} reports its sensitivity.
All experiments are run on NVIDIA A6000 GPUs and averaged over five
seeds (simulated) or three seeds (real).
Appendix~\ref{app:significance} reports paired significance tests
between \realm and \noisy SFT for every configuration.

\subsection{Real Crowd Annotations}
\label{sec:results_real}

Table~\ref{tab:results_real} reports test accuracy on the four
datasets whose annotations come from real crowd workers, with no
simulated noise of any kind.

\textbf{REALM wins under heterogeneous expertise.}
On the three datasets with a wide spread of annotator quality,
\realm is the best method in all twelve cells, across both
annotation regimes and both model sizes, with margins of up to
$7.7$ points.
On WikiAttack it reaches $93.5\%$ against a gold-label oracle ceiling
of $96.3\%$, despite being trained on a pool in which half the workers
agree with the gold label barely more often than chance.
The gains are largest in the \textbf{sparse} regime, where each item
carries a single annotation: this is both the setting a fixed
annotation budget actually buys and the setting in which no
aggregation is possible, so \realm is the only method here
that can exploit annotator identity at all.

\textbf{Aggregation cannot replace annotator modelling.}
In the shared regime, where \mv and \ds become
applicable, majority voting performs at or \emph{below} \noisy
whenever unreliable annotators dominate the vote
($66.6\%$ vs.\ $70.8\%$ on WikiToxicity), confirming that pooling
labels without a quality model can actively discard signal.
\ds is the stronger aggregation baseline and recovers much of
the gap, but it never overtakes \realm in any configuration.
Crucially, both methods collapse the annotator structure into a single
label before training, whereas \realm learns from that
structure and the model jointly, which is why it retains an advantage
even when the aggregation baselines are given every available label.

\textbf{REALM does no harm under homogeneous expertise.}
GoEmotions is annotated by trained raters of uniform quality, and
there all four methods tie to within seed noise
($\Delta$ between $-0.2$ and $+0.4$).
This is what the mechanism predicts rather than a negative
result: \realm's gains track the variance in annotator quality
that is available to exploit, and when that variance vanishes the
mixture reduces to standard fine-tuning.
Together with the preceding results, this delimits when
expertise-aware training is worth its cost.

\subsection{Single-Task Results}
\label{sec:results_uniform}

\begin{table*}[t]
\centering
\caption{Combined test accuracy (\%) on ARC, OBQA, and Riddle in the multi-task setting (mean $\pm$ std over five runs). Oracle accuracy for each model size is shown in parentheses next to the size label.}
\label{tab:results_multi}
\resizebox{0.9\textwidth}{!}{%
\begin{tabular}{|l|l|cc|cc|cc|}
\hline
\multirow{2}{*}{\textbf{Size (Oracle)}} & \multirow{2}{*}{\textbf{Distribution}}
  & \multicolumn{2}{c|}{\textbf{Uniform}}
  & \multicolumn{2}{c|}{\textbf{Asymmetric}}
  & \multicolumn{2}{c|}{\textbf{Systematic}} \\
\cline{3-8}
& & \realm & \noisy
  & \realm & \noisy
  & \realm & \noisy \\
\hline
\multirow{3}{*}{Small (28.57\%)}
  & \dist{1} & $\mathbf{27.89}_{\pm 0.07}$ & $25.20_{\pm 0.69}$ & $\mathbf{27.49}_{\pm 0.32}$ & $25.68_{\pm 0.48}$ & $\mathbf{27.95}_{\pm 0.14}$ & $24.92_{\pm 0.09}$ \\
  & \dist{2} & $\mathbf{28.57}_{\pm 0.21}$ & $28.39_{\pm 0.07}$ & $\mathbf{28.71}_{\pm 0.11}$ & $28.62_{\pm 0.02}$ & $\mathbf{28.39}_{\pm 0.11}$ & $27.93_{\pm 0.11}$ \\
  & \dist{3} & $\mathbf{28.16}_{\pm 0.16}$ & $26.74_{\pm 0.80}$ & $\mathbf{27.98}_{\pm 0.25}$ & $27.01_{\pm 0.07}$ & $\mathbf{28.16}_{\pm 0.11}$ & $26.30_{\pm 0.00}$ \\
\hline
\multirow{3}{*}{Base (46.05\%)}
  & \dist{1} & $\mathbf{42.41}_{\pm 0.67}$ & $38.44_{\pm 0.69}$ & $\mathbf{41.93}_{\pm 0.28}$ & $38.25_{\pm 0.18}$ & $\mathbf{42.05}_{\pm 0.67}$ & $37.75_{\pm 0.55}$ \\
  & \dist{2} & $\mathbf{44.07}_{\pm 0.07}$ & $42.64_{\pm 0.25}$ & $\mathbf{44.18}_{\pm 0.05}$ & $42.48_{\pm 0.05}$ & $\mathbf{43.59}_{\pm 0.09}$ & $42.67_{\pm 0.37}$ \\
  & \dist{3} & $\mathbf{42.51}_{\pm 0.16}$ & $40.30_{\pm 0.48}$ & $\mathbf{42.62}_{\pm 0.09}$ & $41.08_{\pm 0.30}$ & $\mathbf{42.83}_{\pm 0.25}$ & $40.94_{\pm 0.48}$ \\
\hline
\multirow{3}{*}{Large (64.76\%)}
  & \dist{1} & $\mathbf{61.38}_{\pm 0.28}$ & $54.90_{\pm 0.05}$ & $\mathbf{60.44}_{\pm 0.11}$ & $48.69_{\pm 1.75}$ & $\mathbf{60.14}_{\pm 0.00}$ & $53.91_{\pm 0.34}$ \\
  & \dist{2} & $\mathbf{62.41}_{\pm 0.30}$ & $61.49_{\pm 0.07}$ & $\mathbf{62.67}_{\pm 0.32}$ & $60.44_{\pm 0.67}$ & $\mathbf{62.69}_{\pm 0.02}$ & $60.92_{\pm 0.14}$ \\
  & \dist{3} & $\mathbf{62.32}_{\pm 0.11}$ & $59.10_{\pm 0.39}$ & $\mathbf{61.66}_{\pm 0.09}$ & $57.77_{\pm 0.07}$ & $\mathbf{61.08}_{\pm 0.07}$ & $58.46_{\pm 0.71}$ \\
\hline
\end{tabular}}
\end{table*}
\begin{table*}[t]
\centering
\caption{Learned expertise $\hat{\beta}_i = \sigma(\hat{\omega}_i)$,
averaged over five seeds. Each column header is that annotator's
\emph{true} expertise; $\rho$ is the Spearman correlation between true
and learned values for graded pools of $N$ annotators.}
\label{tab:learned_expertise}
\resizebox{0.8\textwidth}{!}{%
\begin{tabular}{|l|ccc|ccc|ccc|cc|}
\hline
\multirow{2}{*}{\textbf{Dataset}}
 & \multicolumn{3}{c|}{\dist{1}}
 & \multicolumn{3}{c|}{\dist{2}}
 & \multicolumn{3}{c|}{\dist{3}}
 & \multicolumn{2}{c|}{\textbf{Spearman} $\rho$} \\
\cline{2-12}
 & $0.9$ & $0.1$ & $0.1$ & $0.9$ & $0.9$ & $0.1$ & $0.9$ & $0.5$ & $0.1$
 & $N{=}5$ & $N{=}10$ \\
\hline
ARC ($K{=}4$)      & 0.96 & 0.26 & 0.25 & 0.92 & 0.94 & 0.11 & 0.95 & 0.71 & 0.21 & 0.94 & 0.89 \\
OBQA ($K{=}4$)     & 0.84 & 0.02 & 0.02 & 0.81 & 0.81 & 0.02 & 0.82 & 0.33 & 0.02 & 1.00 & 0.98 \\
Riddle ($K{=}5$)   & 0.90 & 0.01 & 0.01 & 0.88 & 0.86 & 0.02 & 0.89 & 0.39 & 0.02 & 1.00 & 1.00 \\
PubMedQA ($K{=}3$) & 0.98 & 0.20 & 0.03 & 0.94 & 0.97 & 0.03 & 0.99 & 0.82 & 0.00 & 0.92 & 0.82 \\
PIQA ($K{=}2$)     & 0.28 & 0.00 & 0.00 & 0.44 & 0.44 & 0.01 & 0.30 & 0.01 & 0.00 & 0.98 & 0.84 \\
\hline
\end{tabular}%
}
\end{table*}

\textbf{Results under Uniform Noise.}
Table~\ref{tab:results_uniform} reports test accuracy under uniform
noise across all datasets, model sizes, and expertise distributions.
\realm outperforms \noisy in every setting, and the gain tracks how
much of the training signal is corrupted.
Gains are largest under \dist{1}, where a single reliable
annotator is outnumbered: \noisy absorbs the majority noise directly,
while \realm recovers by down-weighting the two near-random
annotators.
They shrink under \dist{3}, where the intermediate annotator
($\beta{=}0.5$) already supplies a partial signal, and shrink further
under \dist{2}, where two reliable annotators give \noisy a
strong majority to exploit without any expertise weighting.

\textbf{Robustness to Noise Type.}
Table~\ref{tab:results_noise_types} reports \dist{1} across
the three noise types.
\realm stays ahead under all of them, despite asymmetric and
systematic noise violating the uniform-error assumption of the
mixture.
The key insight is that even under misspecified noise the observed
labels still carry information about annotator expertise: unreliable
annotators produce worse labels regardless of the noise structure, so
\realm can still identify and down-weight them.
Asymmetric noise adds a directional bias to the wrong-label
distribution but preserves this signal, and \realm performs
comparably to the uniform setting.
Systematic noise is the hardest case, since a deterministic
per-annotator offset partially masks reliability; the average $\Delta$
falls to $+5.39\%$ (Small) and $+5.13\%$ (Base) from $+6.65\%$ and
$+10.52\%$ under uniform noise, but stays positive throughout.

\textbf{Model Size Effect.}
The $\Delta$ columns of both tables reveal a consistent scaling trend:
the average gain under \dist{1} grows from $+6.65\%$ to
$+10.52\%$ to $+17.25\%$ across Small, Base, and Large under uniform
noise, and from $+7.02\%$ to $+10.50\%$ to $+19.44\%$ under asymmetric
noise.
Larger models have more capacity to turn the expertise estimates into
accuracy.
Under systematic noise the trend flattens, as a partially masked
reliability signal limits what extra capacity can recover.

\textbf{Aggregation baselines.}
\mv and \ds require several labels per example and are undefined in
the sparse regime used here, so we compare against them separately in
the shared regime in Appendix~\ref{app:aggregation}.
\realm remains ahead there: \mv never overtakes it, and \ds, though
competitive early under a favourable annotator pool, degrades with
longer training as it memorises the errors fixed into its aggregated
labels.

% \input{tables/lr_app}

% \textbf{Model Size Effect.}
% The $\Delta$ columns in Tables~\ref{tab:results_uniform}
% and~\ref{tab:results_noise_types} reveal a consistent scaling trend
% across most settings: the average gain under Dist.~\circnum{1} grows
% from $+6.65\%$ to $+10.52\%$ to $+17.25\%$ across Small, Base, and
% Large under uniform noise, and from $+7.02\%$ to $+10.50\%$ to
% $+19.44\%$ under asymmetric noise.
% This suggests that larger models have greater capacity to exploit
% the expertise signal provided by the mixture objective, translating
% annotator reliability estimates into larger accuracy improvements.
% Under systematic noise the trend is less pronounced, likely because
% the deterministic error structure limits how much additional capacity
% can help when the reliability signal itself is partially masked.

\subsection{Multi-Task Results}
\label{sec:results_multi}

Table~\ref{tab:results_multi} reports combined accuracy on ARC, OBQA,
and Riddle in the multi-task setting, where annotator expertise varies
across tasks according to a learned $3{\times}3$ matrix
$\boldsymbol{\Omega}$.
\realm outperforms \noisy in every configuration, across all model
sizes, expertise distributions, and noise types.
As in the single-task setting the largest gains occur under
\dist{1}, where only one annotator is reliable per task:
\realm leads by up to $11.75\%$ (Large, asymmetric: $60.44\%$
vs.\ $48.69\%$), approaching \oracle accuracy in several cases.
Under \dist{2} and \dist{3} the gains are more
modest but remain consistent, indicating that \realm recovers the
structured per-task expertise pattern even when the signal is weaker.
Appendix~\ref{app:per_dataset} breaks the combined figures down by
dataset, where the gains hold on ARC, OBQA, and Riddle individually.

\subsection{Recovering Annotator Expertise}
\label{sec:expertise}

\realm learns one scalar per annotator, $\hat{\beta}_i =
\sigma(\hat{\omega}_i)$, jointly with the model and without ever
seeing a gold label.
Table~\ref{tab:learned_expertise} asks whether those scalars mean
anything.

\textbf{The ranking is recovered everywhere.}
In all fifteen cells the learned ordering matches the true one, and
the Spearman correlation is $0.87$, $0.87$, and $1.00$ across the
three pools.
The value $0.87$ is the \emph{maximum attainable} for the first two
pools, whose true values contain ties that the estimates necessarily
break.
For graded pools of $N{=}5$ and $N{=}10$ the correlation stays between
$0.82$ and $1.00$.
Recovery also survives the asymmetric and systematic channels, and on
the real datasets of Section~\ref{sec:results_real} the learned values
track each worker's measured agreement with gold.

\textbf{Accurate at the top of the range, shrunk at the bottom.}
The estimates are more than an ordering.
On the four multi-class datasets the reliable annotator's expertise is
recovered closely, between $0.81$ and $0.99$ against a true $0.9$.
The error is concentrated at the low end: annotators with true
expertise $0.1$ are driven to between $0.00$ and $0.26$, because the
objective needs only \emph{relative} reliability and labels carrying
no usable signal are pushed toward $\hat{\beta}{=}0$.
Intermediate annotators are ordered correctly but estimated less
sharply, with a true $0.5$ recovered between $0.33$ and $0.82$.
The estimates are therefore informative about how much to trust an
annotator, and most accurate exactly where it matters for training,
though they should not be read as unbiased accuracy estimates at the
unreliable end.
$K{=}2$ is the one case where this breaks down, since the two
components of Equation~\ref{eq:mixture} are closest on binary tasks
and expertise is only weakly identifiable: on PIQA even the reliable
annotator is shrunk to $0.28$, which is also why PIQA requires a large
expertise learning rate (Appendix~\ref{app:lr}).

\begin{table*}[t]
\centering
\caption{Test accuracy (\%) under uniform noise for varying numbers of annotators
(Base model, \dist{3} with uniform expertise spread, mean $\pm$ std over five runs). These runs are re-simulated independently of the main grid.}
\label{tab:ablation_num_annotators}
\resizebox{0.75\textwidth}{!}{%
\begin{tabular}{|l|cc|cc|cc|}
\hline
\multirow{2}{*}{\textbf{Dataset}}
  & \multicolumn{2}{c|}{$N=3$}
  & \multicolumn{2}{c|}{$N=5$}
  & \multicolumn{2}{c|}{$N=10$} \\
\cline{2-7}
& \realm & \noisy
& \realm & \noisy
& \realm & \noisy \\
\hline
ARC      & $\mathbf{45.30}_{\pm 0.26}$ & $43.69_{\pm 0.51}$ & $\mathbf{45.80}_{\pm 0.57}$ & $43.69_{\pm 0.95}$ & $\mathbf{46.25}_{\pm 0.83}$ & $42.70_{\pm 0.60}$ \\
% BioASQ   & $\mathbf{66.67}_{\pm 2.36}$ & $58.54_{\pm 7.88}$ & $\mathbf{63.90}_{\pm 2.22}$ & $57.56_{\pm 6.01}$ & $\mathbf{63.25}_{\pm 2.59}$ & $58.86_{\pm 12.16}$ \\
OBQA     & $\mathbf{57.88}_{\pm 0.24}$ & $55.08_{\pm 0.50}$ & $\mathbf{57.76}_{\pm 0.34}$ & $54.84_{\pm 0.73}$ & $\mathbf{57.80}_{\pm 0.33}$ & $55.00_{\pm 0.44}$ \\
PIQA     & $\mathbf{65.44}_{\pm 0.59}$ & $51.97_{\pm 2.05}$ & $\mathbf{65.70}_{\pm 0.40}$ & $52.03_{\pm 1.28}$ & $\mathbf{65.14}_{\pm 0.63}$ & $51.40_{\pm 0.35}$ \\
PubMedQA & $\mathbf{50.10}_{\pm 1.89}$ & $44.80_{\pm 3.18}$ & $\mathbf{50.35}_{\pm 1.37}$ & $45.00_{\pm 3.28}$ & $\mathbf{47.40}_{\pm 2.34}$ & $47.40_{\pm 2.83}$ \\
Riddle   & $\mathbf{47.80}_{\pm 0.49}$ & $41.33_{\pm 0.72}$ & $\mathbf{47.10}_{\pm 0.73}$ & $41.25_{\pm 1.43}$ & $\mathbf{45.92}_{\pm 0.92}$ & $40.98_{\pm 0.74}$ \\
\hline
\end{tabular}}
\end{table*}

\begin{figure*}[t]
    \centering
    \scriptsize
    \includegraphics[width=0.98\textwidth]{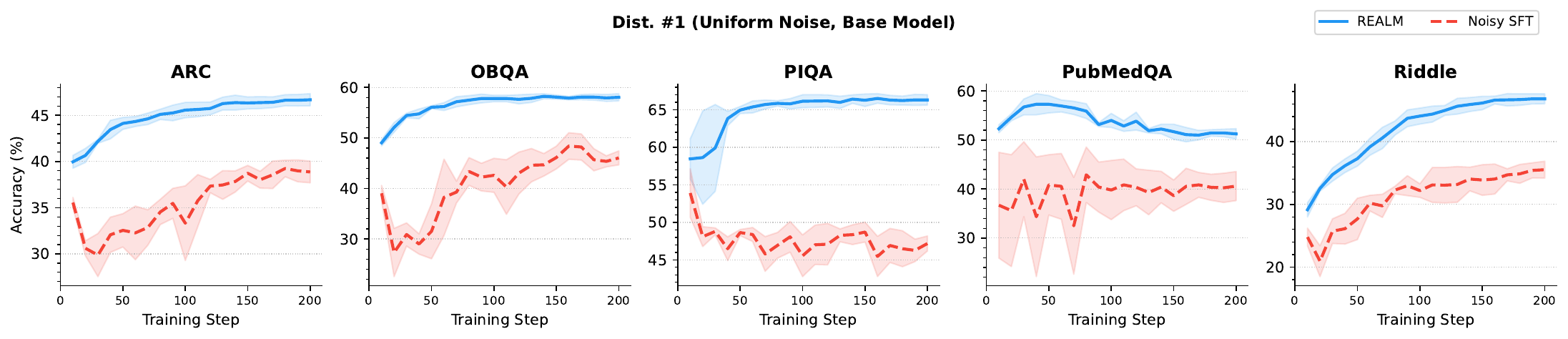}
    \caption{Test accuracy (\%) over training steps under \dist{1} with
    uniform noise, for the Base model across the five datasets.
    Shaded bands denote $\pm$ std over five runs.
    The asymmetric and systematic curves are given in
    Appendix~\ref{app:curves}.}
\label{fig:learning_curves}
\end{figure*}

\begin{table}[t]
\centering
\caption{Accuracy gap $\Delta = \realm - \noisy$ (points) at the
$200$-step budget used throughout the paper and at $5\times$ that
budget, under \dist{1} with uniform noise, base model, five seeds. These runs are re-simulated independently of the main grid.}
\label{tab:training_length}
\resizebox{0.9\columnwidth}{!}{%
\begin{tabular}{|l|ccccc|}
\hline
\textbf{Budget} & ARC & OBQA & PIQA & Riddle & PubMedQA \\
\hline
$200$ steps  & \posdelta{2.0} & \posdelta{3.2} & \posdelta{12.3} & \posdelta{3.1} & \posdelta{6.7} \\
$1000$ steps & \posdelta{3.3} & \posdelta{4.9} & \posdelta{14.3} & \posdelta{6.5} & \posdelta{1.0} \\
\hline
\end{tabular}%
}
\end{table}

% \input{tables/learning_plots}
% \subsection{Ablation Study}

% \textbf{Expertise Learning Rate.}
% Table~\ref{tab:lr_ablation} ablates the expertise scalar learning
% rate $\eta_\omega$ with the model learning rate fixed at
% $5{\times}10^{-5}$.
% Low values ($\eta_\omega \leq 0.05$) cause severe degradation, most
% notably on PIQA ($40.91\%$ at $\eta_\omega{=}0.01$), because the
% expertise scalars adapt too slowly to influence training before the
% model has already absorbed the noise.
% Performance stabilises at $\eta_\omega \geq 0.1$ for most datasets,
% with PIQA requiring $\eta_\omega{=}0.5$ to fully recover, as the
% expertise scalars must adapt quickly enough to influence the model
% before it converges to a noisy solution.

\subsection{Number of Annotators}
\label{sec:num_annotators}

Table~\ref{tab:ablation_num_annotators} reports accuracy as the pool
grows from $N{=}3$ to $N{=}10$ under \dist{3}, with expertise placed
on a uniform grid $\beta_i = 0.9 - i \cdot \tfrac{0.8}{N-1}$ for
$i = 0, \ldots, N-1$.
\realm outperforms \noisy at every dataset and pool size.
Larger $N$ packs more intermediate-reliability annotators into the
same range, so consecutive annotators differ less and are harder to
tell apart; this explains the modest drop in average $\Delta$ from
$+5.93\%$ ($N{=}3$) and $+5.98\%$ ($N{=}5$) to $+5.01\%$ ($N{=}10$).
Repeating the ablation under \dist{1}, where one reliable annotator is
joined by $N-1$ near-random ones and the reliable share falls as
$1/N$, the gap instead \emph{widens} with $N$, reaching $+32.3$ points
on OBQA at $N{=}10$ as \noisy collapses toward chance while \realm
isolates the one informative annotator
(Appendix~\ref{app:annotators_dist1}).

\subsection{Training Dynamics}
\label{sec:curves}

\textbf{Learning curves.}
Figure~\ref{fig:learning_curves} shows test accuracy over training
steps under \dist{1}.
\realm converges smoothly and stays above \noisy throughout, while
\noisy degrades on PIQA as training proceeds and absorbs the dominant
noise from the two unreliable annotators.
\realm also shows noticeably lower seed variance, so jointly learning
annotator expertise stabilises training as well as improving it.
The same pattern holds under the asymmetric and systematic channels
(Appendix~\ref{app:curves}): the separation is widest under asymmetric
noise, where \noisy collapses more severely, and narrowest under
systematic noise, but it is consistent in both.

\textbf{Longer training.}
The $200$-step budget is short in absolute terms, so we retrain for
$5\times$ as long to check that the advantage is not an artefact of
stopping early.
It is not.
Table~\ref{tab:training_length} shows the gap grows on four of the
five datasets and stays positive on the fifth.
The mechanism is the one the method is designed for: with more
training \noisy progressively memorises the corrupted labels
(on ARC it falls from $42.3$ to $40.1$), whereas \realm keeps
down-weighting unreliable annotators and so holds flat or improves.
PubMedQA is the exception, and it is a dataset-size effect rather than
a method effect: its $450$-example split makes $1000$ steps roughly
$140$ epochs, and both methods degrade.
Prolonged training does not destabilise the expertise estimates
either, which continue to sharpen toward their true values at
$2000$ steps (Appendix~\ref{app:impl}).

\section{Conclusion}
We presented \realm, a mixture-based fine-tuning objective that
jointly learns language model parameters and per-annotator expertise
entirely unsupervised.
Modeling each observed label as a convex combination of the model's
prediction and a uniform random guess lets \realm down-weight
unreliable annotators using nothing beyond annotator identity.
On four datasets with real crowdsourced annotations it is the best
method wherever annotator quality is heterogeneous, and the only
applicable one when each item carries a single label.
On five question answering benchmarks with simulated noise it beats
naive noisy SFT in $152$ of $162$ configurations, with gains that grow
with model capacity.
A learned expertise matrix extends the objective to multiple tasks,
and the learned values recover annotator reliability unsupervised.

% Directions for future work include extending REALM to
% state-dependent expertise via learned prompt embeddings, and combining the expertise-aware objective with federated optimization for distributed annotation pipelines.

\section*{Limitations}
\textbf{Fixed expertise per annotator.}
\realm assigns a single scalar expertise value to each annotator,
constant across all prompts.
In practice an annotator may be reliable on certain question types or
domains while performing poorly on others, even within one dataset.
A natural generalisation models expertise as a function of the input,
$\rho_i(Q) = \sigma(\langle \boldsymbol{\omega}_i, \phi(Q) \rangle)$,
where $\phi(Q) \in \mathbb{R}^d$ is a learned prompt embedding and
$\boldsymbol{\omega}_i \in \mathbb{R}^d$ a per-annotator skill vector.
This lets expertise vary continuously across the input space, and
subsumes both the single-task scalar ($d{=}1$) and the multi-task
matrix lookup (when $\phi(Q)$ is the one-hot task encoding) as special
cases.
We leave its full exploration to future work.

\textbf{Fixed label sets.}
The restricted softmax requires a fixed, discrete set of label tokens,
so \realm covers classification and multiple-choice tasks but not
open-ended generation.
Extending it to free-form outputs, where there is no natural $K$-class
space and the uniform component of Equation~\ref{eq:mixture} has no
direct analogue, requires a token-level reference distribution and a
modified mixture objective.

\textbf{Weak identifiability on binary tasks.}
When $K{=}2$ the model component and the uniform component of the
mixture are closest, so expertise is only weakly identified.
Section~\ref{sec:expertise} shows the consequence: on PIQA even the
reliable annotator's expertise is heavily shrunk, and a larger
expertise learning rate is needed for the objective to separate
annotators at all.
More generally, the learned values are accurate for reliable
annotators but biased downward for unreliable ones, so they should be
used to decide how much to trust an annotator rather than as
calibrated estimates of annotation accuracy.

\textbf{Model scale.}
Our experiments use Flan-T5 at up to $780$M parameters.
While the gains grow consistently with capacity across the three sizes
we test, we have not verified that the trend continues at the scale of
contemporary instruction-tuned models, where a stronger prior may
already absorb some annotator noise.

\textbf{Centralised training.}
\realm assumes that all annotations, and the annotator identity
attached to each, are pooled in one place.
Real annotation pipelines often collect data from several independent
organisations under privacy or communication constraints, mirroring
the federated setting where source heterogeneity is a central
challenge.
Parameter-efficient fine-tuning has been extended to federated
\cite{ghiasvand2025communication} and fully decentralized
\cite{ghiasvand2025decentralized} settings with explicit handling of
non-i.i.d.\ data, and combining those with an expertise-aware
objective is a natural direction for future work.

\section*{Acknowledgments}

This work was supported by the National Science Foundation under Grant 2419982,
Grant 2342253, and Grant 2236483.

\bibliography{main.bib}

\clearpage
\appendix

\section{Simulated Annotation Details}
\label{app:simulation}

This appendix gives the full specification of the simulated annotation
protocol summarised in Section~\ref{sec:exp_setup}.
All simulated labels use $N{=}3$ annotators, where annotator $i$
emits the true label with probability $\beta_i$ and otherwise emits an
incorrect label distributed according to the noise type below.

\textbf{Expertise Distributions.}
In the single-task setting we evaluate three annotator expertise
distributions:
\dist{1} ($\beta{=}[0.9, 0.1, 0.1]$), where one
reliable annotator is outnumbered by two near-random ones;
\dist{2} ($\beta{=}[0.9, 0.9, 0.1]$), where two
reliable annotators form a majority; and
\dist{3} ($\beta{=}[0.9, 0.5, 0.1]$), where
expertise is spread uniformly from high to low.
These configurations vary the proportion of reliable annotators, from
a single reliable annotator (\dist{1}) to a majority
(\dist{2}), allowing us to assess \realm across a wide range
of annotation quality regimes.
\dist{1} is the hardest case, since the training signal is
dominated by the two unreliable annotators; \dist{2} is the
easiest, since a naive majority already recovers most of the gold
labels.

\textbf{Noise Types.}
The expertise level $\beta_i$ determines \emph{how often} annotator
$i$ errs; the noise type determines \emph{which} wrong label they
emit.
In the standard \textbf{uniform noise} model assumed by the mixture
objective, when an annotator makes an error the wrong label is drawn
uniformly at random over the $K{-}1$ incorrect classes.
Beyond this, we evaluate two additional noise types that violate this
assumption.
In \textbf{asymmetric noise}, the wrong label concentrates on the
cyclically adjacent class $(t{+}1) \bmod K$, where $t$ is the true
class index, and that class receives a weight of $c{=}4$ times that of
the remaining wrong classes.
Concretely, with $K{=}4$ the adjacent class receives approximately
$57\%$ of the wrong-label mass while the other two incorrect classes
share the remaining $43\%$.
The marginal accuracy $\beta_i$ is preserved, but the wrong-label
distribution is no longer uniform, violating \realm's generative
assumption.
In \textbf{systematic noise}, each annotator $i$ deterministically
emits class $(t + i + 1) \bmod K$ whenever they make an error, so
wrong labels are fully predictable from annotator identity and the
true label.
This is the most severe form of misspecification we consider: the
error channel carries no stochasticity at all, and an annotator's
mistakes are perfectly correlated with their identity, which partially
masks their true reliability.
Together, the three noise types allow us to assess \realm both under
its assumed generative model and under increasingly severe forms of
distributional mismatch.

\textbf{Multi-Task Setup.}
In the multi-task setting we jointly train on three datasets
simultaneously with $N{=}3$ annotators and $M{=}3$ tasks, yielding a
square expertise matrix
$\boldsymbol{\Omega} \in \mathbb{R}^{3 \times 3}$, where each entry
$\Omega_{i,m}$ represents the reliability of annotator $i$ on task
$m$.
We evaluate three structured expertise configurations:
\dist{1}, in which annotator $i$ is reliable only on
task $i$ ($\beta_{i,i}{=}0.9$, $\beta_{i,j}{=}0.1$ for $j \neq i$);
\dist{2}, the inverse pattern in which annotator $i$
is reliable on all tasks except task $i$ ($\beta_{i,i}{=}0.1$,
$\beta_{i,j}{=}0.9$ for $j \neq i$);
and \dist{3}, in which annotator $i$ has expertise
$\beta_{i,j} = [0.9, 0.5, 0.1][(j{-}i) \bmod M]$ on task $j$, so
annotator $i$ is most reliable on their own task $i$ ($\beta{=}0.9$),
intermediate on task $(i{+}1) \bmod M$ ($\beta{=}0.5$), and least
reliable on task $(i{+}2) \bmod M$ ($\beta{=}0.1$).
\dist{1} and \dist{2} are the two extremes of a
sparse versus dense expertise matrix, while \dist{3} produces
a circulant matrix in which every annotator is useful on some task and
useless on another, so no annotator can be globally discarded.

\section{Baseline Details}
\label{app:baselines}

\textbf{Input and target format.}
Every method sees the same training instances and differs only in the
target it is asked to predict and in the loss applied to it.
The input is the question followed by its lettered answer choices, and
the target is a single answer letter.
\noisy SFT takes the annotator's observed label $\tilde{Y}$ as
the target and minimises standard cross-entropy over the restricted
softmax of Equation~\ref{eq:restricted_softmax}, simply dropping the
annotator index $i$ from the training triple $(Q, i, \tilde{Y})$.
\realm consumes the identical triples, retaining $i$ so that the
mixture weight $\sigma(\omega_i)$ of Equation~\ref{eq:loss} can be
applied.
\oracle SFT is trained the same way as \noisy SFT but with the
ground-truth label $Y^{*}$ as the target.
On the real-annotation datasets it reaches $96.3$ (WikiAttack), $74.4$
(WikiToxicity), $90.8$ (SentPolarity), and $70.2$ (GoEmotions) at base
size; these are the ceilings against which
Table~\ref{tab:results_real} should be read.
Model, tokenisation, optimiser, batch size, number of steps, and
random seeds are held identical across all methods, so any difference
in accuracy is attributable to the training objective alone.

\begin{table*}[t]
\centering
\caption{Test accuracy (\%) for varying expertise learning rates $\eta_\omega$ under \dist{1},
uniform noise (Base model, model learning rate $5{\times}10^{-5}$, mean $\pm$ std over five runs).}
\label{tab:lr_ablation}
\resizebox{0.85\textwidth}{!}{%
\begin{tabular}{|l|ccccc|}
\hline
\textbf{Dataset} 
  & ${\text{lr}}=0.01$ 
  & ${\text{lr}}=0.05$ 
  & ${\text{lr}}=0.1$ 
  & ${\text{lr}}=0.5$ 
  & ${\text{lr}}=1.0$ \\
\hline
ARC      & $40.60_{\pm 0.67}$ & $41.96_{\pm 0.54}$ & $42.56_{\pm 0.50}$ & $43.86_{\pm 1.02}$ & $\textbf{43.97}_{\pm 0.77}$ \\
% BioASQ   & $13.01_{\pm 1.78}$ & $24.23_{\pm 21.35}$ & $\mathbf{67.15}_{\pm 0.83}$ & $66.02_{\pm 1.19}$ & $66.50_{\pm 1.58}$ \\
OBQA     & $55.28_{\pm 0.84}$ & $57.68_{\pm 0.37}$ & $57.88_{\pm 0.24}$ & $58.08_{\pm 0.70}$ & $\mathbf{58.24}_{\pm 0.81}$ \\
PIQA     & $40.91_{\pm 3.51}$ & $39.33_{\pm 3.40}$ & $41.85_{\pm 5.80}$ & $\mathbf{65.70}_{\pm 0.67}$ & $65.61_{\pm 0.68}$ \\
PubMedQA & $36.95_{\pm 2.57}$ & $44.15_{\pm 4.67}$ & $49.30_{\pm 3.31}$ & $51.25_{\pm 1.06}$ & $\mathbf{51.80}_{\pm 0.97}$ \\
Riddle   & $42.47_{\pm 2.00}$ & $45.76_{\pm 2.03}$ & $46.04_{\pm 2.07}$ & $46.20_{\pm 2.41}$ & $\mathbf{46.27}_{\pm 2.30}$ \\
\hline
\end{tabular}}
\end{table*}

\textbf{Aggregation baselines.}
\mv{}\,+\,SFT replaces the several labels of each example with
their per-item majority, breaking ties uniformly at random, and then
trains exactly as \noisy SFT on the resulting single-label dataset.
\ds{}\,+\,SFT \cite{dawid1979maximum} estimates a per-annotator
$K \times K$ confusion matrix and a posterior over the true label of
each item by expectation-maximisation, and trains on the hard MAP
label.
Neither has a tunable hyperparameter.
Because both operate on the set of labels attached to an item, they
are defined only in the shared regime; in the sparse regime, where
each example carries a single annotation, majority vote is the
identity map and the Dawid--Skene posterior is unidentifiable, so we
report them as not applicable.

\textbf{Inference.}
Evaluation is identical for every method and involves no annotator
information.
Given a test question with its answer choices, the fine-tuned model
emits an answer token, which is scored against the gold label; the
learned expertise scalars $\boldsymbol{\omega}$ are used only during
training and are discarded at test time.

\section{Training Budgets}
\label{app:impl}

All experiments use an effective batch size of $64$, reached as batch
size $8$ with gradient accumulation over $8$ steps on the simulated
benchmarks, batch size $64$ without accumulation for the
real-annotation base runs, and batch size $16$ with accumulation over
$4$ steps for the real-annotation large runs.
The step counts reported in Section~\ref{sec:exp_setup} are therefore
best read as epoch equivalents, which differ widely across datasets
because the training splits differ widely in size.

\textbf{Simulated-annotation datasets.}
Training runs for $200$ steps, i.e.\ $12{,}800$ training examples.
Relative to the training splits this corresponds to $0.8$ epochs on
PIQA, $2.9$ on OBQA, $3.6$ on RiddleSense, $9.5$ on ARC, and $28$ on
PubMedQA.
The step count is small in absolute terms only because these datasets
are small; in epoch terms it is a standard supervised fine-tuning
budget for classification tasks of this size.
PubMedQA is the outlier in the opposite direction: with only $450$
training examples, even $200$ steps already represent heavy
overtraining, which explains why every method trained on raw labels
degrades on this dataset as the budget grows.
Section~\ref{sec:curves} verifies that \realm's advantage over \noisy
SFT does not depend on this choice, persisting at $5$--$10\times$ the
budget.

\textbf{Real-annotation datasets.}
These have substantially larger training splits and are trained for
$3{,}000$ steps with Flan-T5-base and $2{,}000$ steps with
Flan-T5-large, corresponding to roughly $2$--$6$ epochs on WikiAttack
and WikiToxicity and $7$--$39$ epochs on SentPolarity.
GoEmotions uses $1{,}000$ steps ($2$--$3$ epochs).
Optimisation settings are otherwise identical to the
simulated-annotation experiments, and identical across all methods
compared.

\textbf{Behaviour under prolonged training.}
Because a $200$-step budget invites the objection that the advantage
is an early-training artefact, we also verified that nothing breaks
down at $10\times$ the budget.
Training for $2000$ steps under \dist{1}, the learned expertise of the
near-random annotators remains at zero on all five datasets, while the
reliable annotator's estimate sharpens toward its true value (from
$0.58$ to $1.00$ on ARC).
Accuracy improves rather than degrades over the same interval
(OBQA $56.8$ to $61.6$, Riddle $46.5$ to $57.7$).
The reason is that the mixture down-weights unreliable annotators
early in training, after which their labels contribute vanishing
gradient and are never memorised.

\section{Expertise Learning Rate}
\label{app:lr}

\textbf{Expertise Learning Rate.}
Table~\ref{tab:lr_ablation} ablates the expertise scalar learning
rate $\eta_\omega$ with the model learning rate fixed at
$5{\times}10^{-5}$.
Low values ($\eta_\omega \leq 0.05$) cause severe degradation, most
notably on PIQA ($40.91\%$ at $\eta_\omega{=}0.01$), because the
expertise scalars adapt too slowly to influence training before the
model has already absorbed the noise.
Performance stabilises at $\eta_\omega \geq 0.1$ for most datasets,
with PIQA requiring $\eta_\omega{=}0.5$ to fully recover, as the
expertise scalars must adapt quickly enough to influence the model
before it converges to a noisy solution.

% This ablation is specific to the simulated benchmarks, whose $200$-step
% budget leaves the expertise scalars little time to move.
% The real-annotation experiments train for $1{,}000$--$3{,}000$ steps,
% and there the useful range shifts downward: the values screened,
% $\{0.005, 0.01, 0.1\}$, all train stably, and the selected value was
% never larger than $0.1$.
% What matters is therefore not a large $\eta_\omega$ in absolute terms
% but one large enough relative to the training budget for expertise to
% be identified before the model memorises the noise.

\begin{figure*}[t]
    \centering
    \scriptsize
    \begin{subfigure}[b]{0.98\textwidth}
        \includegraphics[width=\textwidth]{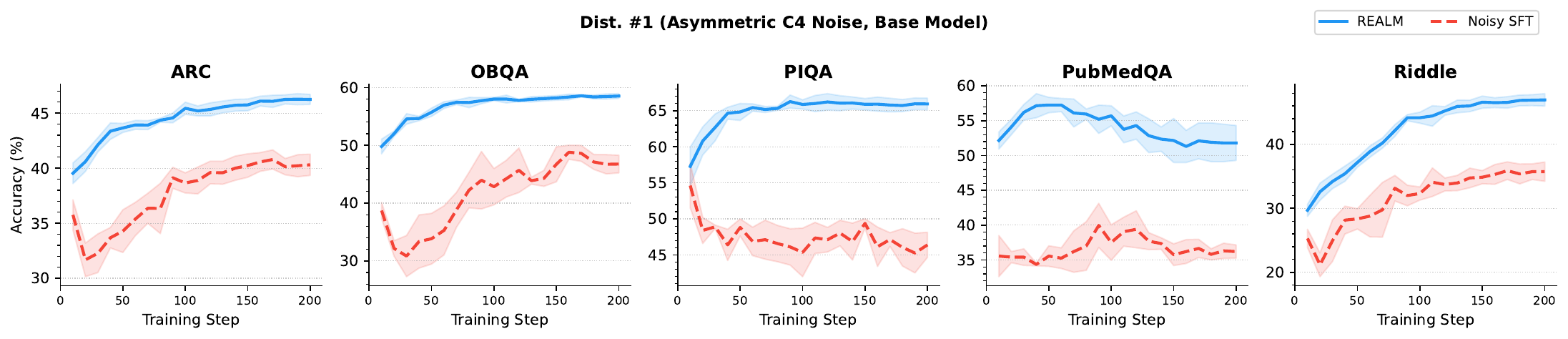}
        \caption{Asymmetric noise.}
    \end{subfigure}
    \begin{subfigure}[b]{0.98\textwidth}
        \includegraphics[width=\textwidth]{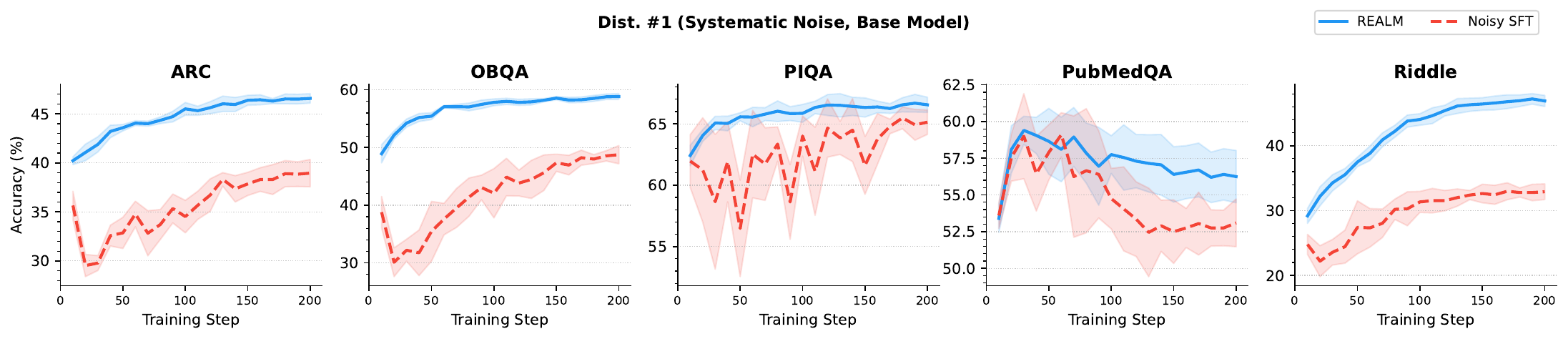}
        \caption{Systematic noise.}
    \end{subfigure}
    \caption{Test accuracy (\%) over training steps under \dist{1} for
    the two noise types that violate the mixture's uniform-error
    assumption, Base model, five datasets. Shaded bands denote $\pm$
    std over five runs.}
\label{fig:learning_curves_app}
\end{figure*}

\section{Aggregation Baselines under Simulation}
\label{app:aggregation}
\begin{table}[t]
\centering
\caption{Aggregation baselines in the \emph{shared} simulated regime
(all three annotators label every example, i.e.\ three times the
annotation budget of our main setting), \dist{1}, uniform noise, base
model, five seeds. $\Delta$ is \realm minus the best baseline.}
\label{tab:agg_sim}
\resizebox{\columnwidth}{!}{%
\begin{tabular}{|l|ccc|cc|}
\hline
\textbf{Dataset} & \noisy & \mv & \ds & \realm & $\Delta$ \\
\hline
ARC      & $35.8$ & $40.9$ & $26.7$ & $\mathbf{44.5}$ & \posdelta{3.6}  \\
OBQA     & $44.9$ & $43.0$ & $19.4$ & $\mathbf{56.6}$ & \posdelta{11.7} \\
PIQA     & $48.1$ & $39.3$ & $35.1$ & $\mathbf{64.8}$ & \posdelta{16.7} \\
Riddle   & $32.1$ & $33.2$ & $19.8$ & $\mathbf{39.4}$ & \posdelta{6.2}  \\
PubMedQA & $35.2$ & $28.5$ & $18.6$ & $\mathbf{60.8}$ & \posdelta{25.6} \\
\hline
\end{tabular}%
}
\end{table}

\begin{table*}[t]
\centering
\caption{\ds's early advantage under \dist{3} is transient. Test
accuracy (\%) at three training budgets, shared regime, base model.
\textbf{Bold} marks the better method at each budget: the advantage
migrates from \ds to \realm as training proceeds, on four of the five
datasets.}
\label{tab:agg_steps}
\resizebox{0.8\textwidth}{!}{%
\begin{tabular}{|l|cc|cc|cc|}
\hline
\multirow{2}{*}{\textbf{Dataset}}
  & \multicolumn{2}{c|}{$200$ steps}
  & \multicolumn{2}{c|}{$1000$ steps}
  & \multicolumn{2}{c|}{$2000$ steps} \\
\cline{2-7}
& \realm & \ds & \realm & \ds & \realm & \ds \\
\hline
ARC      & $45.9$ & $\mathbf{46.3}$ & $\mathbf{46.8}$ & $45.9$ & $\mathbf{46.8}$ & $45.5$ \\
OBQA     & $57.9$ & $\mathbf{59.8}$ & $\mathbf{61.2}$ & $60.8$ & $\mathbf{61.8}$ & $59.8$ \\
PIQA     & $\mathbf{64.3}$ & $51.9$ & $\mathbf{67.7}$ & $50.2$ & $\mathbf{67.8}$ & $50.5$ \\
Riddle   & $46.8$ & $\mathbf{50.9}$ & $57.5$ & $\mathbf{57.6}$ & $\mathbf{58.5}$ & $57.1$ \\
PubMedQA & $59.1$ & $\mathbf{60.4}$ & $51.0$ & $\mathbf{60.4}$ & $52.8$ & $\mathbf{60.6}$ \\
\hline
\end{tabular}%
}
\end{table*}

In our main simulated setting each example carries a single
annotation, so \mv and \ds are undefined there
(Appendix~\ref{app:baselines}).
To compare against them we therefore rerun the simulated experiments
in the shared regime, in which all three annotators label every
example and the aggregation baselines receive three times the
annotation budget available to \realm in the sparse setting.

\textbf{Aggregation is brittle to the composition of the pool.}
Table~\ref{tab:agg_sim} reports the comparison under
\dist{1}, where a single reliable annotator is outnumbered.
\realm is best on all five datasets.
\mv never outperforms \realm in any configuration we ran, and \ds
collapses far below even \noisy, because an unreliable majority drives
its confusion-matrix estimates toward the wrong consensus.
This is the same failure mode observed on real annotations in
Section~\ref{sec:results_real}, where \mv falls below \noisy on
WikiToxicity.

\textbf{\ds's advantage under a favourable pool is transient.}
\ds is a genuinely strong baseline when the pool is more benign: under
the spread expertise of \dist{3} it initially wins on several
of the multi-class datasets.
That advantage does not survive longer training.
Table~\ref{tab:agg_steps} tracks both methods to $2000$ steps: the
model fine-tuned on \ds labels peaks early and then degrades as it
memorises the residual errors baked into the aggregated labels, while
\realm continues to improve and overtakes it on four of the five
datasets.
The exception is PubMedQA, whose $450$-example training split is
heavily overtrained at this budget by any method trained on raw labels
(Appendix~\ref{app:impl}).
The distinction is structural rather than incidental: aggregation
commits to a single hard label \emph{before} training and cannot
revise it, whereas \realm re-estimates annotator reliability jointly
with the model and keeps down-weighting the annotators whose labels
turn out not to fit.

\section{Scaling the Annotator Pool with One Reliable Annotator}
\label{app:annotators_dist1}
\begin{table*}[t]
\centering
\caption{Test accuracy (\%) under uniform noise for varying numbers of annotators
(Base model, \dist{1}: one reliable annotator with $\beta{=}0.9$ and $N-1$
near-random ones with $\beta{=}0.1$, so the share of reliable labels falls as
$1/N$; mean over five runs). Unlike the graded pool of
Table~\ref{tab:ablation_num_annotators}, $\Delta$ \emph{widens} with $N$
on four of the five datasets. These runs are re-simulated independently of the main grid.}
\label{tab:ablation_annotators_dist1}
\resizebox{0.95\textwidth}{!}{%
\begin{tabular}{|l|ccc|ccc|ccc|}
\hline
\multirow{2}{*}{\textbf{Dataset}}
  & \multicolumn{3}{c|}{$N=3$}
  & \multicolumn{3}{c|}{$N=5$}
  & \multicolumn{3}{c|}{$N=10$} \\
\cline{2-10}
& \realm & \noisy & $\Delta$
& \realm & \noisy & $\Delta$
& \realm & \noisy & $\Delta$ \\
\hline
ARC      & $\mathbf{42.6}$ & $40.5$ & \posdelta{2.1}  & $\mathbf{42.8}$ & $37.9$ & \posdelta{4.9}  & $\mathbf{40.8}$ & $32.8$ & \posdelta{8.0}  \\
OBQA     & $\mathbf{57.9}$ & $49.1$ & \posdelta{8.8}  & $\mathbf{55.3}$ & $28.0$ & \posdelta{27.3} & $\mathbf{53.2}$ & $20.9$ & \posdelta{32.3} \\
PIQA     & $\mathbf{65.2}$ & $46.3$ & \posdelta{18.9} & $\mathbf{54.3}$ & $42.4$ & \posdelta{11.9} & $\mathbf{43.6}$ & $39.5$ & \posdelta{4.1}  \\
PubMedQA & $\mathbf{49.4}$ & $40.6$ & \posdelta{8.8}  & $\mathbf{48.4}$ & $27.2$ & \posdelta{21.2} & $\mathbf{42.6}$ & $19.5$ & \posdelta{23.1} \\
Riddle   & $\mathbf{46.0}$ & $36.2$ & \posdelta{9.8}  & $\mathbf{38.7}$ & $25.0$ & \posdelta{13.7} & $\mathbf{34.9}$ & $20.5$ & \posdelta{14.4} \\
\hline
\end{tabular}}
\end{table*}

\begin{table}[t]
\centering
\caption{Paired $t$-test outcomes over the $135$ single-task
configurations ($5$ seeds each). Each row aggregates the $27$
configurations of one dataset.}
\label{tab:significance}
\resizebox{0.8\columnwidth}{!}{%
\begin{tabular}{|l|ccc|}
\hline
\textbf{Dataset} & \realm & \textbf{No sig.} & \noisy \\
 & \textbf{better} & \textbf{diff.} & \textbf{better} \\
\hline
ARC      & 18/27 & 9  & 0 \\
OBQA     & 26/27 & 1  & 0 \\
PIQA     & 15/27 & 11 & 1 \\
PubMedQA & 15/27 & 9  & 3 \\
Riddle   & 25/27 & 2  & 0 \\
\hline
\textbf{All} & \textbf{99/135} & \textbf{32} & \textbf{4} \\
\hline
\end{tabular}
}
\end{table}

\begin{table*}[h]
\centering
\caption{Test accuracy (\%) under \dist{2} across noise types, datasets, and model sizes (mean $\pm$ std over five runs). \textbf{Bold} indicates the better method and
$\Delta = \realm - \noisy$.}
\label{tab:results_dist2_noise_types}
\resizebox{.95\textwidth}{!}{%
\begin{tabular}{|l|l|c|ccc|ccc|ccc|}
\hline
\multirow{2}{*}{\textbf{Size}} & \multirow{2}{*}{\textbf{Dataset}} & \multirow{2}{*}{\textbf{Oracle}}
& \multicolumn{3}{c|}{Uniform}
& \multicolumn{3}{c|}{Asymmetric}
& \multicolumn{3}{c|}{Systematic} \\
\cline{4-12}
& & & \realm & \noisy & $\Delta$
      & \realm & \noisy & $\Delta$
      & \realm & \noisy & $\Delta$ \\
\hline
\multirow{6}{*}{Small}
& ARC      & 29.44 & $\mathbf{29.05}_{\pm 0.47}$ & $27.00_{\pm 0.87}$ & \posdelta{2.05} & $\mathbf{28.70}_{\pm 0.47}$ & $27.54_{\pm 0.57}$ & \posdelta{1.16} & $\mathbf{28.22}_{\pm 0.90}$ & $27.43_{\pm 1.21}$ & \posdelta{0.79} \\
% & BioASQ   & 81.30 & $60.98_{\pm 0.89}$ & $\mathbf{72.36}_{\pm 4.66}$ & \negdelta{11.38} & $61.95_{\pm 1.40}$ & $\mathbf{68.78}_{\pm 4.13}$ & \negdelta{6.83} & $62.60_{\pm 2.06}$ & $\mathbf{70.57}_{\pm 2.79}$ & \negdelta{7.97} \\
& OBQA     & 44.20 & $\mathbf{44.28}_{\pm 0.39}$ & $39.52_{\pm 0.94}$ & \posdelta{4.76} & $\mathbf{44.32}_{\pm 0.50}$ & $40.08_{\pm 0.97}$ & \posdelta{4.24} & $\mathbf{43.68}_{\pm 0.92}$ & $40.00_{\pm 1.67}$ & \posdelta{3.68} \\
& PIQA     & 52.88 & $\mathbf{51.60}_{\pm 1.54}$ & $51.12_{\pm 0.86}$ & \posdelta{0.48} & $50.84_{\pm 1.02}$ & $\mathbf{51.45}_{\pm 1.35}$ & \negdelta{0.61} & $\mathbf{51.32}_{\pm 1.22}$ & $50.84_{\pm 0.92}$ & \posdelta{0.48} \\
& PubMedQA & 58.00 & $\mathbf{53.05}_{\pm 1.88}$ & $50.35_{\pm 3.00}$ & \posdelta{2.70} & $\mathbf{50.95}_{\pm 1.90}$ & $47.05_{\pm 3.75}$ & \posdelta{3.90} & $54.90_{\pm 0.96}$ & $\mathbf{56.50}_{\pm 1.79}$ & \negdelta{1.60} \\
& Riddle   & 39.02 & $\mathbf{34.71}_{\pm 1.14}$ & $31.25_{\pm 1.11}$ & \posdelta{3.46} & $\mathbf{33.49}_{\pm 0.92}$ & $31.76_{\pm 1.41}$ & \posdelta{1.73} & $\mathbf{33.53}_{\pm 0.51}$ & $30.59_{\pm 1.19}$ & \posdelta{2.94} \\
\cline{2-12}
& Avg & 44.71 & $\mathbf{42.54}_{\pm 1.08}$ & $39.85_{\pm 1.36}$ & \textbf{\posdelta{2.69}} & $\mathbf{41.66}_{\pm 0.96}$ & $39.58_{\pm 1.61}$ & \posdelta{2.08} & $\mathbf{42.33}_{\pm 0.90}$ & $41.07_{\pm 1.36}$ & \posdelta{1.26} \\
\hline
\multirow{6}{*}{Base}
& ARC      & 47.47 & $\mathbf{45.49}_{\pm 0.80}$ & $44.31_{\pm 1.01}$ & \posdelta{1.18} & $\mathbf{46.11}_{\pm 0.87}$ & $44.76_{\pm 0.83}$ & \posdelta{1.35} & $\mathbf{45.85}_{\pm 0.34}$ & $44.98_{\pm 0.51}$ & \posdelta{0.87} \\
% & BioASQ   & 87.80 & $66.34_{\pm 2.28}$ & $\mathbf{72.68}_{\pm 3.97}$ & \negdelta{6.34} & $65.37_{\pm 1.32}$ & $\mathbf{72.20}_{\pm 4.49}$ & \negdelta{6.83} & $67.97_{\pm 2.16}$ & $\mathbf{73.17}_{\pm 3.25}$ & \negdelta{5.20} \\
& OBQA     & 59.60 & $\mathbf{58.68}_{\pm 0.81}$ & $56.84_{\pm 0.64}$ & \posdelta{1.84} & $\mathbf{58.24}_{\pm 0.46}$ & $57.08_{\pm 1.16}$ & \posdelta{1.16} & $\mathbf{58.44}_{\pm 0.59}$ & $56.00_{\pm 0.94}$ & \posdelta{2.44} \\
& PIQA     & 67.68 & $\mathbf{66.55}_{\pm 0.52}$ & $63.33_{\pm 1.97}$ & \posdelta{3.22} & $\mathbf{67.18}_{\pm 0.30}$ & $63.55_{\pm 1.79}$ & \posdelta{3.63} & $\mathbf{66.92}_{\pm 0.42}$ & $64.87_{\pm 0.97}$ & \posdelta{2.05} \\
& PubMedQA & 61.00 & $\mathbf{54.00}_{\pm 2.19}$ & $53.20_{\pm 2.15}$ & \posdelta{0.80} & $\mathbf{54.45}_{\pm 1.37}$ & $51.05_{\pm 1.96}$ & \posdelta{3.40} & $56.55_{\pm 1.35}$ & $\mathbf{60.55}_{\pm 1.26}$ & \negdelta{4.00} \\
& Riddle   & 54.12 & $\mathbf{49.61}_{\pm 1.12}$ & $45.84_{\pm 1.03}$ & \posdelta{3.77} & $\mathbf{49.45}_{\pm 0.91}$ & $47.92_{\pm 0.76}$ & \posdelta{1.53} & $\mathbf{49.25}_{\pm 0.89}$ & $46.43_{\pm 1.30}$ & \posdelta{2.82} \\
\cline{2-12}
& Avg & 57.97 & $\mathbf{54.87}_{\pm 1.09}$ & $52.70_{\pm 1.36}$ & \posdelta{2.16} & $\mathbf{55.09}_{\pm 0.78}$ & $52.87_{\pm 1.30}$ & \textbf{\posdelta{2.22}} & $\mathbf{55.40}_{\pm 0.72}$ & $54.57_{\pm 1.00}$ & \posdelta{0.83} \\
\hline
\multirow{6}{*}{Large}
& ARC      & 64.81 & $\mathbf{61.51}_{\pm 0.92}$ & $57.96_{\pm 0.88}$ & \posdelta{3.55} & $\mathbf{62.85}_{\pm 0.92}$ & $59.09_{\pm 0.85}$ & \posdelta{3.76} & $\mathbf{62.40}_{\pm 0.88}$ & $59.64_{\pm 1.16}$ & \posdelta{2.76} \\
% & BioASQ   & 86.99 & $67.48_{\pm 1.54}$ & $\mathbf{76.75}_{\pm 1.51}$ & \negdelta{9.27} & $67.64_{\pm 1.40}$ & $\mathbf{78.86}_{\pm 2.91}$ & \negdelta{11.22} & $67.80_{\pm 1.10}$ & $\mathbf{80.49}_{\pm 3.13}$ & \negdelta{12.69} \\
& OBQA     & 71.60 & $\mathbf{70.68}_{\pm 0.43}$ & $68.80_{\pm 0.46}$ & \posdelta{1.88} & $\mathbf{70.76}_{\pm 0.70}$ & $67.40_{\pm 0.88}$ & \posdelta{3.36} & $\mathbf{71.00}_{\pm 1.07}$ & $69.32_{\pm 0.43}$ & \posdelta{1.68} \\
& PIQA     & 77.80 & $\mathbf{77.67}_{\pm 0.64}$ & $75.10_{\pm 0.23}$ & \posdelta{2.57} & $\mathbf{77.37}_{\pm 0.88}$ & $75.50_{\pm 0.82}$ & \posdelta{1.87} & $\mathbf{77.67}_{\pm 0.63}$ & $75.60_{\pm 1.04}$ & \posdelta{2.07} \\
& PubMedQA & 71.25 & $\mathbf{60.55}_{\pm 1.93}$ & $56.00_{\pm 1.15}$ & \posdelta{4.55} & $\mathbf{61.50}_{\pm 2.93}$ & $56.70_{\pm 3.37}$ & \posdelta{4.80} & $63.80_{\pm 0.64}$ & $\mathbf{67.75}_{\pm 1.26}$ & \negdelta{3.95} \\
& Riddle   & 67.84 & $\mathbf{65.45}_{\pm 0.57}$ & $61.49_{\pm 1.24}$ & \posdelta{3.96} & $\mathbf{64.43}_{\pm 0.73}$ & $62.04_{\pm 1.56}$ & \posdelta{2.39} & $\mathbf{65.45}_{\pm 0.73}$ & $62.75_{\pm 0.57}$ & \posdelta{2.70} \\
\cline{2-12}
& Avg & 70.66 & $\mathbf{67.17}_{\pm 0.90}$ & $63.87_{\pm 0.79}$ & \textbf{\posdelta{3.30}} & $\mathbf{67.38}_{\pm 1.23}$ & $64.15_{\pm 1.50}$ & \posdelta{3.23} & $\mathbf{68.06}_{\pm 0.79}$ & $67.01_{\pm 0.89}$ & \posdelta{1.05} \\
\hline
\end{tabular}%
}
\end{table*}

\begin{table*}[t]
\centering
\caption{Test accuracy (\%) under \dist{3} across noise types, datasets, and model sizes (mean $\pm$ std over five runs). \textbf{Bold} indicates the better method and
$\Delta = \realm - \noisy$.}
\label{tab:results_dist3_noise_types}
\resizebox{0.95\textwidth}{!}{%
\begin{tabular}{|l|l|c|ccc|ccc|ccc|}
\hline
\multirow{2}{*}{\textbf{Size}} & \multirow{2}{*}{\textbf{Dataset}} & \multirow{2}{*}{\textbf{Oracle}}
& \multicolumn{3}{c|}{Uniform}
& \multicolumn{3}{c|}{Asymmetric}
& \multicolumn{3}{c|}{Systematic} \\
\cline{4-12}
& & & \realm & \noisy & $\Delta$
      & \realm & \noisy & $\Delta$
      & \realm & \noisy & $\Delta$ \\
\hline
\multirow{6}{*}{Small}
& ARC      & 29.44 & $\mathbf{26.70}_{\pm 0.51}$ & $26.64_{\pm 1.40}$ & \posdelta{0.06} & $27.21_{\pm 0.96}$ & $\mathbf{27.43}_{\pm 0.96}$ & \negdelta{0.22} & $\mathbf{27.45}_{\pm 1.28}$ & $26.76_{\pm 0.75}$ & \posdelta{0.69} \\
% & BioASQ   & 81.30 & $\mathbf{53.98}_{\pm 2.22}$ & $51.38_{\pm 7.97}$ & \posdelta{2.60} & $\mathbf{53.50}_{\pm 3.83}$ & $49.27_{\pm 6.27}$ & \posdelta{4.23} & $62.60_{\pm 2.06}$ & $\mathbf{70.57}_{\pm 2.79}$ & \negdelta{7.97} \\
& OBQA     & 44.20 & $\mathbf{42.52}_{\pm 0.45}$ & $35.80_{\pm 2.04}$ & \posdelta{6.72} & $\mathbf{42.48}_{\pm 0.82}$ & $35.76_{\pm 1.88}$ & \posdelta{6.72} & $\mathbf{42.56}_{\pm 1.11}$ & $35.00_{\pm 1.77}$ & \posdelta{7.56} \\
& PIQA     & 52.88 & $\mathbf{52.88}_{\pm 1.18}$ & $50.36_{\pm 0.60}$ & \posdelta{2.52} & $50.25_{\pm 0.87}$ & $\mathbf{50.58}_{\pm 0.39}$ & \negdelta{0.33} & $\mathbf{51.32}_{\pm 1.22}$ & $50.84_{\pm 0.92}$ & \posdelta{0.48} \\
& PubMedQA & 58.00 & $\mathbf{50.60}_{\pm 1.97}$ & $46.55_{\pm 3.17}$ & \posdelta{4.05} & $\mathbf{49.00}_{\pm 3.44}$ & $36.80_{\pm 3.08}$ & \posdelta{12.20} & $53.80_{\pm 0.94}$ & $\mathbf{55.00}_{\pm 0.74}$ & \negdelta{1.20} \\
& Riddle   & 39.02 & $\mathbf{32.16}_{\pm 1.44}$ & $29.69_{\pm 0.69}$ & \posdelta{2.47} & $\mathbf{32.59}_{\pm 1.41}$ & $29.10_{\pm 0.80}$ & \posdelta{3.49} & $\mathbf{32.27}_{\pm 1.14}$ & $28.16_{\pm 1.64}$ & \posdelta{4.11} \\
\cline{2-12}
& Avg & 44.71 & $40.97_{\pm 1.11}$ & $37.81_{\pm 1.58}$ & \posdelta{3.16} & $40.31_{\pm 1.50}$ & $35.93_{\pm 1.42}$ & \textbf{\posdelta{4.38}} & $41.48_{\pm 1.14}$ & $39.15_{\pm 1.16}$ & \posdelta{2.33} \\
\hline
\multirow{6}{*}{Base}
& ARC      & 47.47 & $\mathbf{43.78}_{\pm 0.60}$ & $42.15_{\pm 0.39}$ & \posdelta{1.63} & $\mathbf{44.00}_{\pm 0.86}$ & $42.76_{\pm 1.07}$ & \posdelta{1.24} & $\mathbf{43.90}_{\pm 0.47}$ & $43.52_{\pm 0.68}$ & \posdelta{0.38} \\
% & BioASQ   & 87.80 & $\mathbf{66.99}_{\pm 2.09}$ & $58.54_{\pm 7.88}$ & \posdelta{8.45} & $\mathbf{66.83}_{\pm 2.26}$ & $53.33_{\pm 11.10}$ & \posdelta{13.50} & $67.97_{\pm 2.16}$ & $\mathbf{73.17}_{\pm 3.25}$ & \negdelta{5.20} \\
& OBQA     & 59.60 & $\mathbf{57.96}_{\pm 0.54}$ & $55.00_{\pm 1.28}$ & \posdelta{2.96} & $\mathbf{57.72}_{\pm 1.11}$ & $55.12_{\pm 0.77}$ & \posdelta{2.60} & $\mathbf{58.60}_{\pm 0.95}$ & $54.36_{\pm 1.24}$ & \posdelta{4.24} \\
& PIQA     & 67.68 & $\mathbf{65.79}_{\pm 0.78}$ & $51.47_{\pm 1.23}$ & \posdelta{14.32} & $\mathbf{65.77}_{\pm 0.71}$ & $52.10_{\pm 1.75}$ & \posdelta{13.67} & $\mathbf{66.92}_{\pm 0.42}$ & $64.87_{\pm 0.97}$ & \posdelta{2.05} \\
& PubMedQA & 61.00 & $\mathbf{49.15}_{\pm 1.85}$ & $44.80_{\pm 3.18}$ & \posdelta{4.35} & $\mathbf{52.00}_{\pm 2.42}$ & $43.70_{\pm 3.32}$ & \posdelta{8.30} & $55.85_{\pm 1.93}$ & $\mathbf{58.45}_{\pm 1.81}$ & \negdelta{2.60} \\
& Riddle   & 54.12 & $\mathbf{46.63}_{\pm 2.01}$ & $42.00_{\pm 1.88}$ & \posdelta{4.63} & $\mathbf{46.94}_{\pm 1.34}$ & $42.47_{\pm 1.17}$ & \posdelta{4.47} & $\mathbf{46.47}_{\pm 1.31}$ & $42.08_{\pm 0.95}$ & \posdelta{4.39} \\
\cline{2-12}
& Avg & 57.97 & $52.66_{\pm 1.16}$ & $47.08_{\pm 1.59}$ & \posdelta{5.58} & $53.29_{\pm 1.29}$ & $47.23_{\pm 1.62}$ & \textbf{\posdelta{6.06}} & $54.35_{\pm 1.02}$ & $52.66_{\pm 1.13}$ & \posdelta{1.69} \\
\hline
\multirow{6}{*}{Large}
& ARC      & 64.81 & $\mathbf{59.59}_{\pm 0.64}$ & $56.14_{\pm 1.07}$ & \posdelta{3.45} & $\mathbf{59.64}_{\pm 1.08}$ & $55.93_{\pm 0.47}$ & \posdelta{3.71} & $\mathbf{58.49}_{\pm 0.59}$ & $56.74_{\pm 0.54}$ & \posdelta{1.75} \\
% & BioASQ   & 86.99 & $\mathbf{67.32}_{\pm 2.02}$ & $60.16_{\pm 8.46}$ & \posdelta{7.16} & $\mathbf{66.50}_{\pm 1.58}$ & $55.93_{\pm 14.57}$ & \posdelta{10.57} & $67.80_{\pm 1.10}$ & $\mathbf{80.49}_{\pm 3.13}$ & \negdelta{12.69} \\
& OBQA     & 71.60 & $\mathbf{70.12}_{\pm 1.24}$ & $65.68_{\pm 0.52}$ & \posdelta{4.44} & $\mathbf{69.80}_{\pm 1.21}$ & $64.80_{\pm 1.03}$ & \posdelta{5.00} & $\mathbf{70.12}_{\pm 1.03}$ & $65.40_{\pm 1.08}$ & \posdelta{4.72} \\
& PIQA     & 77.80 & $\mathbf{76.50}_{\pm 0.28}$ & $51.99_{\pm 3.98}$ & \posdelta{24.51} & $\mathbf{76.17}_{\pm 0.49}$ & $53.47_{\pm 2.70}$ & \posdelta{22.70} & $\mathbf{77.67}_{\pm 0.63}$ & $75.60_{\pm 1.04}$ & \posdelta{2.07} \\
& PubMedQA & 71.25 & $\mathbf{53.45}_{\pm 2.09}$ & $48.05_{\pm 3.08}$ & \posdelta{5.40} & $\mathbf{55.70}_{\pm 2.32}$ & $47.45_{\pm 2.70}$ & \posdelta{8.25} & $60.95_{\pm 2.11}$ & $\mathbf{64.05}_{\pm 2.85}$ & \negdelta{3.10} \\
& Riddle   & 67.84 & $\mathbf{62.31}_{\pm 1.65}$ & $57.06_{\pm 2.38}$ & \posdelta{5.25} & $\mathbf{61.33}_{\pm 0.96}$ & $56.16_{\pm 0.98}$ & \posdelta{5.17} & $\mathbf{61.25}_{\pm 0.97}$ & $57.92_{\pm 1.91}$ & \posdelta{3.33} \\
\cline{2-12}
& Avg & 70.66 & $64.39_{\pm 1.18}$ & $55.78_{\pm 2.21}$ & \posdelta{8.61} & $64.53_{\pm 1.21}$ & $55.56_{\pm 1.58}$ & \textbf{\posdelta{8.97}} & $65.70_{\pm 1.07}$ & $63.94_{\pm 1.48}$ & \posdelta{1.76} \\
\hline
\end{tabular}%
}
\end{table*}

Table~\ref{tab:ablation_annotators_dist1} repeats the annotator-count
ablation of Section~\ref{sec:num_annotators} under \dist{1}, the
hardest version of the experiment: a single reliable annotator is
joined by $N-1$ near-random ones, so the share of reliable labels
falls as $1/N$ rather than staying fixed.
\realm leads at every pool size on every dataset, and unlike the
graded pool the margin \emph{widens} with $N$ on four of the five
datasets: \noisy degrades toward chance as the near-random majority
grows, while \realm continues to concentrate weight on the one
informative annotator.
On OBQA the gap grows from $+8.8$ points at $N{=}3$ to $+32.3$ at
$N{=}10$; on PubMedQA, from $+8.8$ to $+23.1$.
PIQA is the exception, where the gap narrows from $+18.9$ to $+4.1$
because \realm itself degrades with $N$ ($65.2$ to $43.6$).
This is the $K{=}2$ identifiability limit of
Section~\ref{sec:expertise}: on a binary task the mixture separates
the reliable annotator from the rest only weakly, so a shrinking
reliable share hurts \realm as well.
The learned expertise confirms the mechanism directly: on ARC, OBQA,
and Riddle the reliable annotator receives the highest $\hat{\beta}$
on five of five seeds at every $N$, reaching $0.74$ against $0.004$
for the others on OBQA at $N{=}10$.

\section{Learning Curves under Misspecified Noise}
\label{app:curves}

Figure~\ref{fig:learning_curves_app} gives the training curves under
the two noise channels that violate the mixture's uniform-error
assumption, complementing the uniform-noise curves of
Figure~\ref{fig:learning_curves}.
\realm remains above \noisy throughout training in both, and retains
the lower seed variance seen under uniform noise.

\section{Significance Testing}
\label{app:significance}

\textbf{Protocol.}
A \emph{configuration} is one cell of our experimental grid, defined
by a dataset, model size, noise type, and expertise distribution, and
trained with five random seeds.
Within each configuration, \realm and \noisy SFT are trained on
\emph{identical} annotations and paired by seed, so the only source of
variation between the two arms is the training objective.
We apply a two-sided paired $t$-test to the five per-seed accuracy
differences and report significance at $p < 0.05$.

We use the paired $t$-test rather than the Wilcoxon signed-rank test
because with five paired samples the two-sided Wilcoxon $p$-value
cannot fall below $0.0625$, and so can never reach the conventional
threshold.
As a nonparametric check we instead report the sign pattern: in $93$
of the $103$ significant configurations, the winning method is ahead
on all five seeds, which is the strongest outcome the Wilcoxon test
can register at this sample size.

\textbf{Results.}
Table~\ref{tab:significance} aggregates the outcome over all $135$
single-task configurations, with each row summarising the $27$
configurations of one dataset ($3$ model sizes $\times$ $3$ noise
types $\times$ $3$ expertise distributions).
\realm is significantly better in $99$ of $135$ configurations and
significantly worse in $4$.
The remaining $32$ show no significant difference at $p < 0.05$.
The four reversals are concentrated on PubMedQA and PIQA, the two
datasets with the smallest training splits and the smallest label
spaces respectively.

\begin{table*}[t]
\centering
\caption{Per-dataset test accuracy (\%) in the multi-task setting across model sizes,
expertise distributions, and noise types (mean $\pm$ std over five runs).
\textbf{Bold} indicates the better method within each configuration.
This table provides the per-dataset breakdown of the combined results
reported in Table~\ref{tab:results_multi}.}
\label{tab:multi_per_dataset}
\resizebox{0.94\textwidth}{!}{%
\begin{tabular}{|l|l|l|cc|cc|cc|}
\hline
\multirow{2}{*}{\textbf{Size}} & \multirow{2}{*}{\textbf{Distribution}} & \multirow{2}{*}{\textbf{Dataset}}
  & \multicolumn{2}{c|}{\textbf{Uniform}}
  & \multicolumn{2}{c|}{\textbf{Asymmetric}}
  & \multicolumn{2}{c|}{\textbf{Systematic}} \\
\cline{4-9}
& & & \realm & \noisy
    & \realm & \noisy
    & \realm & \noisy \\
\hline
\multirow{9}{*}{Small}
  & \multirow{3}{*}{\dist{1}}
    & ARC    & $\mathbf{26.35}_{\pm 0.34}$ & $25.54_{\pm 0.47}$ & $25.92_{\pm 0.94}$ & $\mathbf{26.39}_{\pm 0.73}$ & $\mathbf{26.61}_{\pm 0.86}$ & $26.05_{\pm 0.21}$ \\
  & & OBQA   & $\mathbf{34.90}_{\pm 0.70}$ & $29.50_{\pm 1.50}$ & $\mathbf{34.60}_{\pm 1.40}$ & $29.50_{\pm 0.50}$ & $\mathbf{35.80}_{\pm 1.40}$ & $26.20_{\pm 0.40}$ \\
  & & Riddle & $\mathbf{24.51}_{\pm 1.76}$ & $20.20_{\pm 0.39}$ & $\mathbf{24.12}_{\pm 0.59}$ & $20.29_{\pm 0.10}$ & $\mathbf{23.33}_{\pm 0.00}$ & $21.08_{\pm 0.29}$ \\
\cline{2-9}
  & \multirow{3}{*}{\dist{2}}
    & ARC    & $26.39_{\pm 0.04}$ & $\mathbf{27.42}_{\pm 0.04}$ & $26.78_{\pm 0.09}$ & $\mathbf{27.77}_{\pm 0.21}$ & $26.52_{\pm 0.52}$ & $\mathbf{27.04}_{\pm 0.43}$ \\
  & & OBQA   & $\mathbf{38.10}_{\pm 1.10}$ & $36.90_{\pm 0.30}$ & $\mathbf{37.20}_{\pm 0.80}$ & $35.40_{\pm 0.40}$ & $\mathbf{36.20}_{\pm 1.00}$ & $34.90_{\pm 0.30}$ \\
  & & Riddle & $\mathbf{24.22}_{\pm 0.10}$ & $22.25_{\pm 0.10}$ & $\mathbf{24.80}_{\pm 0.10}$ & $23.92_{\pm 0.20}$ & $\mathbf{25.00}_{\pm 0.29}$ & $23.14_{\pm 0.78}$ \\
\cline{2-9}
  & \multirow{3}{*}{\dist{3}}
    & ARC    & $26.22_{\pm 0.64}$ & $\mathbf{26.48}_{\pm 0.39}$ & $\mathbf{26.35}_{\pm 0.77}$ & $26.22_{\pm 0.64}$ & $\mathbf{26.70}_{\pm 0.77}$ & $26.35_{\pm 0.00}$ \\
  & & OBQA   & $\mathbf{35.90}_{\pm 1.50}$ & $33.40_{\pm 2.40}$ & $\mathbf{34.80}_{\pm 0.80}$ & $34.60_{\pm 1.20}$ & $\mathbf{35.50}_{\pm 1.10}$ & $30.70_{\pm 0.30}$ \\
  & & Riddle & $\mathbf{25.00}_{\pm 0.69}$ & $20.78_{\pm 0.20}$ & $\mathbf{25.00}_{\pm 0.10}$ & $21.37_{\pm 0.00}$ & $\mathbf{24.31}_{\pm 0.20}$ & $21.86_{\pm 0.29}$ \\
\hline
\multirow{9}{*}{Base}
  & \multirow{3}{*}{\dist{1}}
    & ARC    & $\mathbf{41.76}_{\pm 1.07}$ & $38.67_{\pm 0.90}$ & $\mathbf{42.15}_{\pm 0.52}$ & $40.26_{\pm 0.34}$ & $\mathbf{42.19}_{\pm 0.73}$ & $38.76_{\pm 0.99}$ \\
  & & OBQA   & $\mathbf{55.00}_{\pm 0.40}$ & $47.90_{\pm 1.10}$ & $\mathbf{54.30}_{\pm 0.30}$ & $45.90_{\pm 0.70}$ & $\mathbf{54.20}_{\pm 0.20}$ & $47.70_{\pm 0.10}$ \\
  & & Riddle & $\mathbf{31.57}_{\pm 0.78}$ & $28.63_{\pm 0.20}$ & $\mathbf{29.31}_{\pm 0.29}$ & $26.18_{\pm 0.69}$ & $\mathbf{29.80}_{\pm 0.98}$ & $25.69_{\pm 0.20}$ \\
\cline{2-9}
  & \multirow{3}{*}{\dist{2}}
    & ARC    & $\mathbf{42.53}_{\pm 0.04}$ & $41.93_{\pm 0.64}$ & $\mathbf{42.79}_{\pm 0.30}$ & $42.19_{\pm 0.30}$ & $\mathbf{42.32}_{\pm 0.00}$ & $41.72_{\pm 0.60}$ \\
  & & OBQA   & $\mathbf{55.50}_{\pm 0.70}$ & $55.30_{\pm 0.90}$ & $\mathbf{55.80}_{\pm 0.20}$ & $53.50_{\pm 1.30}$ & $\mathbf{54.90}_{\pm 0.30}$ & $54.80_{\pm 1.60}$ \\
  & & Riddle & $\mathbf{36.37}_{\pm 0.49}$ & $31.86_{\pm 0.49}$ & $\mathbf{35.98}_{\pm 0.69}$ & $32.35_{\pm 0.78}$ & $\mathbf{35.39}_{\pm 0.69}$ & $32.94_{\pm 1.76}$ \\
\cline{2-9}
  & \multirow{3}{*}{\dist{3}}
    & ARC    & $\mathbf{41.37}_{\pm 0.60}$ & $40.52_{\pm 0.34}$ & $\mathbf{42.49}_{\pm 0.26}$ & $41.50_{\pm 0.47}$ & $\mathbf{42.49}_{\pm 0.34}$ & $41.29_{\pm 0.77}$ \\
  & & OBQA   & $\mathbf{54.90}_{\pm 0.70}$ & $52.30_{\pm 2.10}$ & $\mathbf{54.30}_{\pm 0.70}$ & $53.20_{\pm 0.40}$ & $\mathbf{55.00}_{\pm 0.00}$ & $52.80_{\pm 0.40}$ \\
  & & Riddle & $\mathbf{32.94}_{\pm 0.00}$ & $28.04_{\pm 0.78}$ & $\mathbf{31.47}_{\pm 0.29}$ & $28.24_{\pm 0.59}$ & $\mathbf{31.67}_{\pm 0.29}$ & $28.53_{\pm 0.69}$ \\
\hline
\multirow{9}{*}{Large}
  & \multirow{3}{*}{\dist{1}}
    & ARC    & $\mathbf{62.10}_{\pm 0.04}$ & $58.76_{\pm 0.56}$ & $\mathbf{61.24}_{\pm 0.21}$ & $51.89_{\pm 1.59}$ & $\mathbf{61.67}_{\pm 0.04}$ & $57.73_{\pm 0.73}$ \\
  & & OBQA   & $\mathbf{68.40}_{\pm 0.20}$ & $60.90_{\pm 0.90}$ & $\mathbf{68.10}_{\pm 0.90}$ & $52.20_{\pm 2.20}$ & $\mathbf{67.30}_{\pm 1.10}$ & $61.30_{\pm 0.90}$ \\
  & & Riddle & $\mathbf{52.84}_{\pm 1.47}$ & $40.20_{\pm 1.96}$ & $\mathbf{51.08}_{\pm 0.88}$ & $37.94_{\pm 1.67}$ & $\mathbf{49.61}_{\pm 0.98}$ & $37.94_{\pm 1.08}$ \\
\cline{2-9}
  & \multirow{3}{*}{\dist{2}}
    & ARC    & $\mathbf{62.06}_{\pm 0.52}$ & $61.89_{\pm 0.77}$ & $\mathbf{62.62}_{\pm 0.13}$ & $61.29_{\pm 0.86}$ & $\mathbf{62.45}_{\pm 0.04}$ & $61.03_{\pm 0.09}$ \\
  & & OBQA   & $\mathbf{69.50}_{\pm 0.30}$ & $67.80_{\pm 1.20}$ & $\mathbf{70.40}_{\pm 0.40}$ & $66.10_{\pm 1.10}$ & $\mathbf{70.40}_{\pm 0.20}$ & $67.70_{\pm 0.10}$ \\
  & & Riddle & $\mathbf{56.27}_{\pm 0.39}$ & $54.41_{\pm 0.88}$ & $\mathbf{55.20}_{\pm 0.69}$ & $52.94_{\pm 0.20}$ & $\mathbf{55.69}_{\pm 0.39}$ & $54.02_{\pm 0.29}$ \\
\cline{2-9}
  & \multirow{3}{*}{\dist{3}}
    & ARC    & $\mathbf{62.83}_{\pm 0.26}$ & $60.94_{\pm 0.26}$ & $\mathbf{62.10}_{\pm 0.04}$ & $58.97_{\pm 0.26}$ & $\mathbf{61.93}_{\pm 0.04}$ & $60.82_{\pm 0.04}$ \\
  & & OBQA   & $\mathbf{69.40}_{\pm 1.00}$ & $65.20_{\pm 0.40}$ & $\mathbf{68.70}_{\pm 0.30}$ & $65.40_{\pm 0.20}$ & $\mathbf{68.00}_{\pm 0.40}$ & $65.60_{\pm 1.40}$ \\
  & & Riddle & $\mathbf{54.22}_{\pm 1.08}$ & $48.92_{\pm 1.47}$ & $\mathbf{53.73}_{\pm 0.78}$ & $47.55_{\pm 0.69}$ & $\mathbf{52.35}_{\pm 0.59}$ & $46.08_{\pm 1.76}$ \\
\hline
\end{tabular}}
\end{table*}

\textbf{Win counts.}
Significance aside, it is also useful to count outright wins by mean
accuracy, since this is the summary quoted in the abstract.
The full grid comprises $162$ configurations: the $135$ single-task
cells above plus $27$ multi-task cells ($3$ model sizes $\times$ $3$
noise types $\times$ $3$ expertise distributions).
\realm attains a higher five-seed mean than \noisy in $152$ of them.
The ten exceptions are exactly the negative $\Delta$ entries of
Tables~\ref{tab:results_noise_types},
\ref{tab:results_dist2_noise_types},
and~\ref{tab:results_dist3_noise_types}: seven are PubMedQA, whose
$450$-example training split leaves every method unstable at this
budget, two are PIQA, the only binary task, and one is ARC.
No reversal occurs under uniform noise, and none in the multi-task
setting, whose configurations are scored by the combined accuracy of
Table~\ref{tab:results_multi}.

The test outcomes also confirm the capacity trend reported in
Section~\ref{sec:results_uniform}: the number of significant wins
grows monotonically with model size, from $23$ (Small) to $37$ (Base)
to $39$ (Large) out of $45$ configurations per model size.
Larger models are better able to convert the expertise signal supplied
by the mixture objective into accuracy gains that are detectable above
seed variance.

\section{Additional Results}

% \subsection{Effect of Annotation Mode}
% \label{sec:annotation_mode}

% \input{tables/plot_exclusive_small_app}
% \input{tables/plot_exclusive_base_app}
% \input{tables/plot_exclusive_large_app}

% Figure~\ref{fig:bar_comparison} compares shared and exclusive annotation
% modes for Base and Large models under uniform noise.
% In the shared mode every annotator labels every example, tripling the
% training set size but also the proportion of noisy labels; in the
% exclusive mode each annotator labels a disjoint partition, preserving
% the original dataset size with a lower but still heterogeneous noise
% level per example.

% \realm outperforms \noisy under both modes and across
% both distributions, confirming that the gains reported in the preceding
% tables are not an artifact of the exclusive setting.
% Under \dist{1}, the shared mode amplifies the damage from unreliable
% annotators — \noisy (shared) collapses on BioASQ and PIQA —
% while \realm (shared) recovers substantially, in some cases
% exceeding \realm (exclusive).
% Under \dist{3}, where annotation quality is higher, the two modes
% converge and the differences between \realm and \noisy
% are smaller but consistent across datasets.

\subsection{Results under the Remaining Expertise Distributions}
Tables~\ref{tab:results_dist2_noise_types} and~\ref{tab:results_dist3_noise_types}
report results under the remaining two expertise distributions.
Under \dist{2}, where two reliable annotators form a majority,
gains are modest (avg $\Delta \leq +4\%$) since \noisy can
already exploit the strong majority signal, with PubMedQA reversing
under systematic noise at Base and Large model sizes. Under \dist{3}, gains are larger and asymmetric noise yields
the highest average $\Delta$ across all model sizes, while systematic
noise again produces the smallest gains with occasional reversals on
PubMedQA. Overall, \realm outperforms \noisy in the large majority of settings across both distributions, noise types, and
model sizes.

% \clearpage

\subsection{Per-Dataset Multi-Task Results}
\label{app:per_dataset}

Table~\ref{tab:multi_per_dataset} provides the per-dataset breakdown
of the combined multi-task results reported in Table~\ref{tab:results_multi}.
\realm outperforms \noisy on the large majority of
individual dataset and configuration combinations, confirming that the
gains observed in the combined accuracy are not driven by a single
dataset. The most consistent and largest gains appear on OBQA and Riddle,
which have more answer choices ($K{=}4$ and $K{=}5$ respectively),
giving the mixture objective a stronger signal to discriminate reliable
from unreliable annotators. ARC shows smaller but still consistent gains, its four answer choices
giving the mixture less to work with than Riddle's five.
The five exceptions in the table are all ARC at the Small model size,
where absolute accuracy sits close to the $25\%$ chance level and the
two methods are separated by at most $1.1$ points.
None of them surfaces in the combined accuracy of
Table~\ref{tab:results_multi}, which \realm leads in all $27$
multi-task configurations.

% \clearpage
% \input{src/future}

\end{document}